\documentclass{article}

\PassOptionsToPackage{numbers, compress}{natbib}
\usepackage[preprint]{neurips_2026}

\usepackage[utf8]{inputenc}
\usepackage[T1]{fontenc}
\usepackage{hyperref}
\usepackage{url}
\usepackage{booktabs}
\usepackage{amsfonts}
\usepackage{amsmath}
\usepackage{amssymb}
\usepackage[capitalize,noabbrev]{cleveref}
\usepackage{pifont}
\newcommand{\cmark}{\ding{51}}
\newcommand{\xmark}{\ding{55}}
\usepackage{nicefrac}
\usepackage{microtype}
\usepackage[table]{xcolor}
\usepackage{graphicx}

\definecolor{better1}{RGB}{230,240,250}  % p < 0.05
\definecolor{better2}{RGB}{200,220,245}  % p < 0.01
\definecolor{better3}{RGB}{170,200,235}  % p < 0.001
\definecolor{worse1}{RGB}{250,235,235}   % p < 0.05
\definecolor{worse2}{RGB}{245,215,215}   % p < 0.01
\definecolor{worse3}{RGB}{235,195,195}   % p < 0.001
\usepackage{subcaption}
\usepackage{algorithm}
\usepackage{algorithmic}
\usepackage{multirow}
\usepackage{wrapfig}
\usepackage{placeins}
\usepackage{tikz}
\usetikzlibrary{positioning, arrows.meta, calc, fit, backgrounds}

\newcommand{\method}{ScreenShot}

\title{\method{}: A Foundation Model for Few-Shot Combination Drug Screening}
\author{%
  Antoine de Mathelin, Christopher Tosh and Wesley Tansey \\
  Computational Oncology\\
  Memorial Sloan Kettering Cancer Center\\
  New York, NY 10065 \\
  \texttt{tanseyw@mskcc.org} \\
}

\begin{document}
\maketitle

% ======================================================================
% ======================================================================
\begin{abstract}
Treating patients with combinations of drugs reduces the risk of resistance to any individual drug. Finding effective combinations is difficult because the large search space makes combinatorial screens prohibitively expensive, time consuming, and often technically infeasible. Predictive models can fill this gap, yet existing methods typically require molecular profiling of each sample and per-cohort training, limiting their applicability when time and tissue are scarce. To address this challenge, we introduce \method{}, a hierarchical transformer pretrained on 40 drug screening datasets covering 3,700 drugs and 6,000 biological samples, whose architecture mirrors the nested structure of screening data. Given a few-shot context of observations from a new patient, \method{} predicts the response of the sample to combination therapies through in-context learning, operating directly on functional measurements with no fine-tuning and no molecular profiling. On four held-out datasets, \method{} outperforms all baselines in both prediction accuracy and identification of selectively effective treatments. \method{}'s internal representations are directly useful for experimental design: we use them to drive a weighted $k$-means++ active learning strategy that selects which experiments to run, achieving the same hit detection as uniform screening with a third of the budget. Source code and interactive dashboard: \url{https://github.com/tansey-lab/screenshot}.~\looseness=-1
\end{abstract}

% ======================================================================
\section{Introduction}
\label{sec:intro}
% ======================================================================
Drug screens enable functional characterization of tumors in terms of vulnerabilities to candidate therapies. Screens in triple negative breast cancer~\cite{guillen2022human}, fibrolamellar carcinoma~\cite{lalazar:etal:2021:fibrolamellar-carcinoma-pdc}, glioblastoma~\cite{lee:etal:2018:pancan-pdc,johansson:etal:2020:gbm-pdc}, pancreatic adenocarcinoma~\cite{tiriac:etal:2018:pancreas-pdo,driehuis:etal:2019:panc-pdo,hirt:etal:2022:pdac-pdo}, acute myeloid leukemia~\cite{lee:etal:2023:aml-pdcs,burd:etal:2020:beat-aml}, and epithelial ovarian cancers~\cite{sa:etal:2019:gyn-pdc,martins:etal:2022:hgsoc-pdc,murumagi:etal:2022:ovarian-pdc} have revealed novel vulnerabilities and new potential therapies for patients. Multiple recent clinical trials~\cite{peterziel:etal:2022:peditaric-pdc,lau:etal:2021:pediatric-pdx-pdc} have integrated drug screens into the clinical decision making process and demonstrated that screen-guided treatments can improve patient outcomes. ~\looseness=-1

A key limitation of existing screens and trials is the inability to scale the number of experiments. In many clinical settings, only a small number of cells can be gathered, limiting the number of experiments possible. Further, all trials to-date have focused on single drug screens with combinations designed based on ad hoc assessments of putative drug targets and the perceived molecular susceptibility of a tumor to combination targeting. Similarly, the largest combination screens to-date~\cite{oneil:etal:2016:merck-cl-combo,holbeck:etal:2019:nci-almanac-combo,jaaks2022effective} screened only a small subset of combinations, since exhaustively enumerating all combinations would have been cost and time prohibitive. There is a pressing need to develop methods for combination drug screens that maximize sample efficiency and can determine responses to a broad panel of drugs based on relatively few observations.~\looseness=-1

% Ex vivo drug screening---testing drugs directly on a patient's tumor cells---offers a direct route to personalized cancer treatment~\citep{sachs2018living, vlachogiannis2018patient, wensink2021patient}. In principle, measuring viability across many drugs, doses, and combinations could identify effective therapies for each individual. In practice, the search space is combinatorial: even 100 drugs screened pairwise across a few dose levels requires millions of measurements per patient, far beyond what limited material and clinical timelines allow~\citep{tosh2025batchie, peterziel:etal:2022:inform-trial}.

% This bottleneck is now addressable because decades of ex vivo screening studies have been consolidated into a pan-cancer atlas spanning 40+ datasets~\citep{pichotta2026pan}. We use this corpus to pretrain \method{}, a transformer-based foundation model for drug screening.

To address this need, we developed \method{}, a transformer-based foundation model for few-shot prediction in combination drug screens. \method{} performs \emph{in-context learning} at inference: given a small set of drug-dose-viability measurements from a new patient sample, it predicts the remaining (untested) responses, without fine-tuning and without requiring molecular profiling of the sample. This contrasts with prior methods that rely on genomic or transcriptomic features to encode biological context~\citep{kuenzi2020predictingDLDR, liu2020deepcdr, jin2021hidra}, which are often unavailable or inconsistent for patient-derived samples~\citep{pichotta2026pan}. By operating directly on the functional measurements produced by the screen, \method{} can be applied to new samples with no feature engineering and without the need for time and cost-intensive molecular profiling.~\looseness=-1

Beyond prediction, we develop a multi-round active learning strategy to identify the most effective treatments with minimal experimental budget.
The strategy selects experiments using $k$-means++ seeding~\citep{arthur2007k} in \method{}'s embedding space, weighting candidates by their predicted effectiveness to prioritize likely hits while ensuring diversity across rounds.
In our benchmarks, our adaptive selection algorithm achieves the same hit recall as uniform selection with only a third of the budget. Our contributions are:~\looseness=-1
\begin{itemize}
    \item \textbf{Foundation model for functional drug response.} We introduce \method{}, a hierarchical transformer pretrained on a pan-cancer collection of 40 drug screen datasets, which predicts dose-response behavior for new patient samples via in-context learning using only drug-dose-viability measurements (no omics features; no per-patient finetuning).
    \item \textbf{Embedding-driven experimental design.} We develop a practical screening policy that uses \method{}'s pretrained drug-dose embeddings for cold-start batch selection and its context-conditioned embeddings for active learning, explicitly trading off testing promising treatments with broad coverage of the drug-dose design space.
    \item \textbf{Strong empirical performance.} We demonstrate state-of-the-art few-shot prediction and hit detection on held-out monotherapy and combination screens, outperforming XGBoost~\cite{chen:guestrin:2016:xgboost}, TabPFN~\cite{hollmann2025accurate}, and pretrained neural network baselines across budgets.
\end{itemize}

% ======================================================================

% ======================================================================
\section{Method}
\label{sec:method}
% ======================================================================

\subsection{Problem formulation}
\label{sec:formulation}

We consider a drug screen setting in which a biological sample $s$ (e.g., a cell line or organoid) is treated with a drug combination $\mathbf d=(d_1,\ldots,d_q)$ at respective doses $\boldsymbol\delta=(\delta_1,\ldots,\delta_q)$, producing a viability measurement $y\in[0,1]$. The combination size $q$ varies across experiments (e.g., $q{=}1$ for monotherapy, $q{=}2$ for pairwise combinations, etc.). We assume a fixed drug library of size $M$, and encode each drug identity as an integer index $d\in\{1,\ldots,M\}$; doses are real-valued, $\delta\in\mathbb R$. Given a small context set of labeled observations for a new sample $s$, $\mathcal D_s=\{(\mathbf d_j,\boldsymbol\delta_j,y_j)\}_{j=1}^{n}$, our goal is to predict the viability for untested drug-dose queries $(\mathbf d',\boldsymbol\delta')$ on the same sample, outputting a prediction $\hat y$. To best match many precision medicine settings where only small biopsies and thus limited cells are available, we assume that no molecular profiling of $s$ (e.g., gene expression or mutation data) is available: the model must condition only on the observed (drug, dose, viability) tuples in $\mathcal D_s$.~\looseness=-1

Formally, ScreenShot is a parametric predictor $f_\theta$ with parameters $\theta \in \mathbb{R}^W$ that maps a query perturbation and its context to a viability estimate,
\[
f_\theta:\underbrace{\big(\{1,\ldots,M\}^q \times \mathbb R^q\big)}_{\text{query}}\times\underbrace{\big(\{1,\ldots,M\}^q \times \mathbb R^q \times [0,1]\big)^{n}}_{\text{few-shot context}}\to[0,1],
\qquad
\hat{y}=f_\theta\!\big(\underbrace{(\mathbf d',\boldsymbol\delta')}_{\text{query}},\,\underbrace{\mathcal D_s}_{\text{context}}\big).
\]
This differs from the classical transfer-learning approach, where one first trains a global predictor $h_\theta:\{1,\ldots,M\}^q \times \mathbb R^q \to [0,1]$ and then fine-tunes it on $\mathcal D_s$ to obtain sample-specific parameters $\theta_s$, yielding predictions $\hat{y}=h_{\theta_s}(\mathbf d',\boldsymbol\delta')$. In contrast, \method{} follows the paradigm of recent foundation models for tabular data such as TabPFN~\citep{hollmann2025accurate}, and amortizes the adaptation step by conditioning directly on $\mathcal D_s$ via cross-attention, effectively replacing explicit fine-tuning with in-context inference. This design is practical and computationally efficient, and avoids the hyperparameter-sensitive fine-tuning trade-offs between fitting $\mathcal D_s$ and preserving prior knowledge (e.g., learning rate, number of epochs, and regularization)~\citep{howard2018universal, mosbach2021stability}.~\looseness=-1

\subsection{Model architecture}
\label{sec:architecture}

\begin{figure}[htbp]
    \centering
    \includegraphics[width=\textwidth]{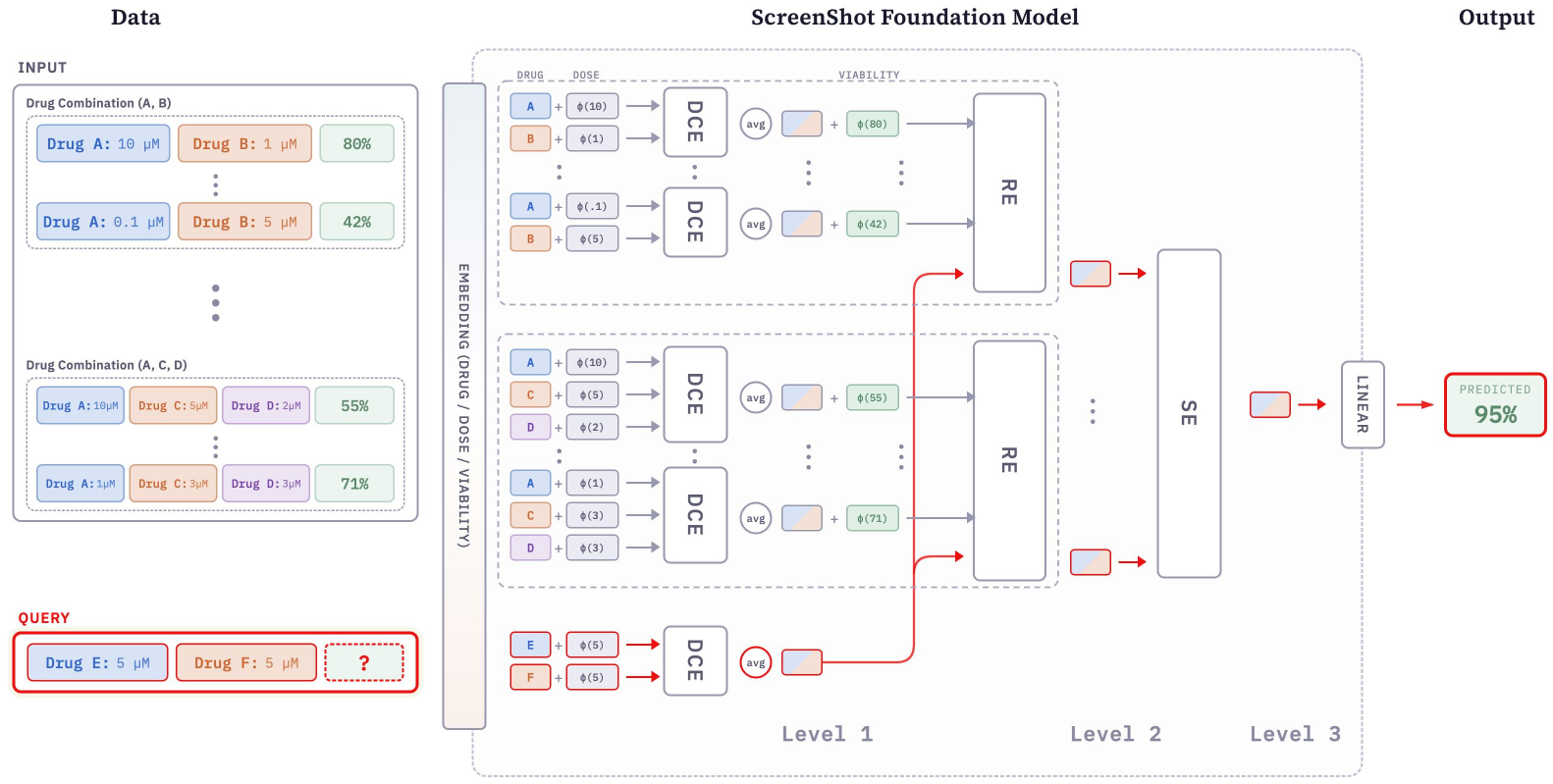}
    \caption{\textbf{Overview of the \method{} architecture.} \method{} processes drug screening data through a three-level hierarchy. Level~1: all drug + dose embeddings are combined via self-attention within each drug combination by the \textit{Drug Combination Encoder} (DCE). Level~2: query combinations attend to context observations (augmented with observed viability) via cross-attention through the \textit{Response Encoder} (RE). Level~3: perturbation-level embeddings are aggregated via self-attention across the sample by the \textit{Sample Encoder} (SE), producing a final representation for viability prediction.}
    \label{fig:architecture}
\end{figure}

\method{} is a hierarchical transformer that leverages the compositional structure of drug screening data (\cref{fig:architecture}). The key observation is that measurements for a fixed drug combination across different dose levels naturally form a group (a dose-response curve) and that each combination is itself composed of individual drugs. Rather than representing the entire context as a single flat sequence of observations, \method{} organizes it into three nested levels: \textit{Level 1}: drug-dose combination (perturbation), \textit{Level 2}: response to the perturbation, \textit{Level 3}: aggregation of all perturbations across the sample. An analogy in natural language processing is to first encode words into sentence representations and then attend across sentences in a document, instead of attending over all words in the document at once~\citep{yang2016hierarchical}. For a target query $(\mathbf d',\boldsymbol\delta')$, \method{} performs cross-attention between the query representation and the context at Level~2, enabling it to retrieve and aggregate evidence from dose-response curves of similar perturbations. All three encoders at the three levels use the same transformer configuration: 5 attention blocks, 8 heads, GeLU activations.~\looseness=-1
  
The hierarchical architecture is motivated by both modeling and practical considerations. First, the hierarchy mirrors the data generating structure and yields representations at multiple granularities (individual drug, combination, and sample-level) that can be reused in downstream tasks. Second, it reduces the computational cost of attention. A na\"ive flat-attention model over all $n$ context observations would require $O(n^2)$ attention computations. In contrast, the hierarchical design factorizes this cost into: (i) cross-attention from the query to context observations at Level 2, $\mathcal{O}(n)$, and (ii) self-attention across perturbations, $O(p^2)$, where $p$ is the number of unique perturbations in the context. For a typical screen with $n{=}300$ observations comprising $p{=}50$ unique perturbations at $6$ dose points each, the dominant attention cost decreases from $300^2{=}90{,}000$ to $50^2{=}2{,}500$ attention entries. This shares the same computational motivation as efficient attention methods that mitigate the quadratic cost via local or block-sparse patterns~\citep{child2019generating, zaheer2020big}; however, in our case the partition is induced by the assay structure rather than an arbitrary chunking of the input sequence.~\looseness=-1

\paragraph{Drug, dose, and viability embeddings.}
Each drug identity $d_i$ in the library is mapped to a learned embedding $\mathbf e_{d_i}\in\mathbb R^{D}$ via an embedding table. Crucially, these representations are learned purely from \emph{function}, namely how drugs affect viability across diverse samples, doses, and combination partners, rather than from chemical structure. This makes \method{} agnostic to drug modality (e.g., small molecules, natural products, antibodies) and more flexible than approaches based on molecular descriptors, which may be incomplete or unavailable for certain compound classes. For novel compounds not present in the pretraining corpus, the model uses dedicated \textsc{unk} embeddings that are learned jointly during pretraining and allow the model to rely on in-context observations to infer the drug's behavior (see \cref{app:drug_coverage}).~\looseness=-1

Doses and viabilities are encoded via learned Fourier features $\phi(\cdot)$ that map continuous scalars to $\mathbb{R}^D$ (details in \cref{app:architecture}). We use $D{=}256$ throughout.~\looseness=-1

\paragraph{Drug combination encoder (Level 1).}
The drug embedding and dose encoding are summed, $\mathbf{h}_k = \mathbf{e}_{d_k} + \phi(\delta_k)$, and the resulting sequence $(\mathbf{h}_1, \ldots, \mathbf{h}_q)$ is processed by a transformer encoder. Self-attention allows each drug token to attend to the other drugs in the combination, capturing interaction effects. The outputs are aggregated via masked average pooling into a single combination embedding $\mathbf{c} \in \mathbb{R}^D$. Because the number of drugs $q$ is handled through attention masking, the same architecture natively supports monotherapies, pairs, and higher-order combinations without any structural modification. This encoder is applied to each context perturbation $(\mathbf d_j,\boldsymbol\delta_j)$, yielding a sequence of embeddings $(\mathbf c_1,\ldots,\mathbf c_n)$, and is likewise applied to the query $(\mathbf d',\boldsymbol\delta')$ to produce a query embedding $\mathbf c'$.~\looseness=-1

\paragraph{Response encoder (Level 2).}
At this level, \method{} conditions the query embedding $\mathbf c'$ on the sample-specific context using multi-head cross-attention. Each context observation is represented by augmenting its combination embedding with the measured viability,
$\tilde{\mathbf c}_j = \mathbf c_j + \phi_y(y_j)$.
The context is then grouped by unique perturbation (i.e., all dose points measured for the same drug combination). Cross-attention is applied separately within each group: the query $\mathbf c'$ attends to the observations in perturbation group $g$, producing a group-conditioned query representation $\mathbf r'_g$. Repeating this over all groups produces $(\mathbf r'_1,\ldots,\mathbf r'_p)$, one representation per observed perturbation. When predicting multiple queries for the same sample, this procedure is applied independently to each query, so that predictions depend only on the shared context and the corresponding query.~\looseness=-1

\paragraph{Sample encoder (Level~3).}
\method{} then aggregates information across the perturbations observed for the same sample. The sequence $(\mathbf r'_1,\ldots,\mathbf r'_p)$ is processed by a transformer encoder with multi-head self-attention, allowing evidence to propagate across perturbations (e.g., response patterns for one drug informing predictions for related drugs). The resulting query representation is mapped to a viability estimate via a linear prediction head followed by a sigmoid, producing $\hat y\in[0,1]$.~\looseness=-1

% \subsection{Multi-level embeddings}
% \label{sec:embeddings}

% A distinctive property of the hierarchical architecture is that it produces embeddings at multiple levels of abstraction, each serving different downstream purposes (\cref{tab:embedding_levels}).
% The drug embeddings (Level~0) capture intrinsic pharmacological properties and can be used for drug similarity analysis and transfer to other tasks.
% The drug-dose combination embeddings (Level~1) represent treatments in a structured space that does not require any viability observations, making them ideal for \emph{cold-start} experimental design: by clustering in this space, we can select a diverse initial batch of experiments before any patient data is available (\S\ref{sec:active_learning}).
% The context-conditioned embeddings (Levels~2--3) incorporate observed viability patterns and can be used for \emph{adaptive} experimental design, where the model's current understanding of the sample guides the selection of follow-up experiments.

\subsection{Pretraining}
\label{sec:pretraining}

\method{} is pretrained on a pan-cancer drug screening atlas~\cite{pichotta2026pan} spanning both monotherapy and combination assays across diverse screening technologies and biological systems.
Overall, the corpus contains approximately 30M viability measurements across about 4k drugs and 6k biological samples, including cancer cell lines as well as patient-derived models such as ex vivo cultures and organoids.~\looseness=-1

Training follows an in-context learning objective that matches the downstream setting.
Let $\mathcal{S}_{\text{train}}$ denote the set of biological samples available during pretraining, where each sample $s \in \mathcal{S}_{\text{train}}$ comes with a set of labeled observations $\mathcal{D}_s$.
At each training step, we sample a biological sample $s \sim \mathcal{S}_{\text{train}}$, then draw a context set $\mathcal{D}\subset \mathcal{D}_s$ and a query set $\mathcal{Q}\subset \mathcal{D}_s$ uniformly at random, with elements of the form $(\mathbf d,\boldsymbol\delta,y)$.
We train \method{} to predict the query viabilities conditioned on the context,
$\hat{y}=f_\theta\!\big((\mathbf d,\boldsymbol\delta),\,\mathcal{D}\big)$ for each $(\mathbf d,\boldsymbol\delta,y)\in\mathcal{Q}$.

To handle technical noise (e.g. failed experiments) in the data, we optimize $\theta$ by minimizing the mean absolute error over the query set,
\[
\mathcal{L}(\theta) = \mathbb{E}_{s,\mathcal{D},\mathcal{Q}}\!\left[\frac{1}{|\mathcal{Q}|}\sum_{(\mathbf d,\boldsymbol\delta,y)\in \mathcal{Q}} \bigl|f_\theta\!\big((\mathbf d,\boldsymbol\delta),\,\mathcal{D}\big) - y\bigr|\right],
\]
which is more robust to outliers than the mean squared error.~\looseness=-1
No molecular profiling features are used at any stage; the model is trained only from drug identity, dose, and observed viability in the context.
Implementation details are provided in \cref{app:architecture,app:datasets}.

\subsection{Few-shot inference}
\label{sec:fewshot}

At inference time, \method{} is applied to a previously unseen sample $s \sim \mathcal{S}_{\text{test}}$, drawn from a held-out cohort.
Given a small context set of labeled measurements $\mathcal{D}_s$, the model is queried on untested drug-dose conditions.
\method{} conditions on $\mathcal{D}_s$ through cross-attention and produces predictions $\hat{y}$ for each query. No gradient updates or fine-tuning are performed: the model generalizes to new samples entirely through its pretrained attention mechanism, which learns to condition on the provided context. This is analogous to in-context learning in large language models, where a pretrained transformer adapts its predictions based on the input prompt without updating its weights~\citep{brown2020gpt3, hollmann2025accurate}.~\looseness=-1

\subsection{Active learning pipeline}
\label{sec:active_learning}

\method{} enables few-shot screening: given a small set of (drug, dose, viability) observations from a new sample, it predicts the remaining responses through in-context inference.
The practical bottleneck is therefore deciding \emph{which} experiments to measure.
Because \method{}'s prediction accuracy depends on the context it receives, we seek a budgeted strategy for selecting context points that helps the model identify the most effective and selective treatments as quickly as possible. We use a multi-round design in which we select a batch of treatments to measure, append the new observations to the context, and re-run \method{} to update predictions and representations.~\looseness=-1

\paragraph{Cold-start batch.}
The first batch must be chosen before any viability is observed for the target sample, which is the cold-start setting~\citep{de2025addressing}.
At this stage, we rely on representations that do not require labels.
We compute Level 1 drug-dose embeddings for all candidate treatments and select an initial batch using $k$-medoids clustering~\citep{kaufman1990partitioning}.
This targets coverage in the pretrained treatment space, providing a diverse initial context that is informative on average across samples.~\looseness=-1

\paragraph{Adaptive batches.}
After observing the initial measurements, \method{} produces context-conditioned representations for each candidate treatment.
While these representations could be used with generic rules such as coverage ($k$-medoids) or uncertainty-based acquisition, we found it more effective to drive selection by the downstream objective: identifying selective hits.
For each unobserved treatment $t$, we compute predicted differential viability
$\Delta_t = \hat{y}_t^{(\mathrm{sample})} - \hat{y}_t^{(\mathrm{control})}$,
where the control term is defined as in \cref{sec:experiments}.
Strongly negative $\Delta_t$ indicates predicted selectivity. To select the next batch, we run weighted $k$-means++ seeding~\citep{arthur2007k} on the context-conditioned treatment representations.
Candidates are weighted by $w_t = \max(-\Delta_t, 0)$ to prioritize predicted hits.
Treatments already selected in previous rounds are included as fixed initial centers, so that new selections are pushed away from already-observed regions of the treatment space.
We compare this objective-driven policy to coverage-only and uncertainty-based alternatives in \cref{app:ablation}; we find it recovers top hits more efficiently under the same budget.~\looseness=-1

\begin{algorithm}[t]
\caption{Multi-round active learning with \method{}}
\label{alg:active_learning}
\begin{algorithmic}[1]
\REQUIRE Unlabeled pool $\mathcal{U}$, cold-start budget $n_0$, budget per round $b$, rounds $T$
\STATE Compute drug-dose embeddings $\mathbf{c}_t$ for all $t \in \mathcal{U}$ \hfill $\triangleright$ Level 1 (DCE)
\STATE $\mathcal{D}_s \leftarrow \text{$k$-medoids}(\{\mathbf{c}_t\}, n_0)$; and obtain labels for $\mathcal{D}_s$ \hfill $\triangleright$ Cold-start
\FOR{$r = 1, \ldots, T$}
    \STATE Predict $\hat{y}_t$ for all $t \in \mathcal{U}$ using \method{} conditioned on $\mathcal{D}_s$
    \STATE $\Delta_t = \hat{y}_{\text{target},t} - \min_{c \in \mathcal{C}} \hat{y}_{c,t}$ for each $t \notin \mathcal{D}_s$
    \STATE $w_t = \max(-\Delta_t, 0)$ \hfill $\triangleright$ Prioritize predicted hits
    \STATE Extract embeddings $\mathbf{h}_t$ for all $t \in \mathcal{U}$ \hfill $\triangleright$ Level 3 (SE)
    \STATE $\mathcal{B} \leftarrow \text{$k$-means++}(\{\mathbf{h}_t\}_{t \notin \mathcal{D}_s},\; b,\; \{w_t\},\; \text{init}{=}\{\mathbf{h}_t\}_{t \in \mathcal{D}_s})$; and obtain labels for $\mathcal{B}$
    \STATE $\mathcal{D}_s \leftarrow \mathcal{D}_s \cup \mathcal{B}$
\ENDFOR
\RETURN Final predictions from \method{} conditioned on $\mathcal{D}_s$
\end{algorithmic}
\end{algorithm}

% ======================================================================

% ======================================================================
\section{Related work}
\label{sec:related}
% ======================================================================

\paragraph{Drug response prediction.}
Machine learning for drug response prediction has been extensively studied, with deep learning methods proposed for monotherapy sensitivity~\citep{sakellaropoulos2019deepDLDR, liu2019tcnns, kuenzi2020predictingDLDR, liu2020deepcdr, jin2021hidra, nguyen2022graphdrp} and combination synergy~\citep{preuer2018deepsynergy, wang2022deepdds, rafiei2023deeptrasynergy, kuru2022matchmaker, elkhili2023marsy, julkunen2020comboltr}. A pervasive limitation is the reliance on genomic or transcriptomic features to represent biological samples, which restricts applicability to patient-derived samples where such profiling is often unavailable or inconsistent~\citep{pichotta2026pan}. Transfer learning approaches~\citep{ma2021fewshot, sederman2024screendlTLDR, mourragui2021predictingTLDR} partially address the data scarcity problem by pretraining on cell line databases, but still require overlapping molecular features between source and target domains. \method{} removes this requirement entirely, operating only on drug-dose-viability measurements.~\looseness=-1

\paragraph{Foundation models and in-context learning.}
Foundation models pretrained on large corpora and adapted through in-context learning, without fine-tuning on the target task, have achieved strong few-shot performance in natural language processing~\citep{brown2020gpt3} and, more recently, on tabular data~\citep{hollmann2025accurate, kim2024carte, ma2024tabdpt}. This paradigm replaces the fragile fine-tuning step, which is sensitive to learning rate, regularization, and dataset size~\citep{howard2018universal, mosbach2021stability}, with a single forward pass conditioned on a context set. \method{} adopts this paradigm for drug screening: it is pretrained on a pan-cancer atlas and performs in-context inference on new patients. Several concurrent works explore foundation models or in-context learning for drug response prediction~\citep{liu2025building, li2024cancergpt, edwardssynergpt, pichotta2026pan}. However, these approaches present significant limitations. Some rely on knowledge encoded in large language models, requiring prior literature on the sample, which is unavailable for patient-derived cells~\citep{li2024cancergpt, liu2025building}. Others predict binary synergy labels rather than continuous dose-response, a strictly less flexible setting since dose-response models can derive synergy but not the reverse~\citep{li2024cancergpt, edwardssynergpt}. All process context observations through flat attention, scaling quadratically in context size. Most do not release pretrained weights~\citep{li2024cancergpt, edwardssynergpt, pichotta2026pan}, and~\citet{pichotta2026pan} is limited to monotherapy. \method{} addresses all of these limitations.~\looseness=-1

\paragraph{Active learning for drug screening.}
Active learning has shown promise for drug screening~\citep{reker2015activeLdrugdisco, graff2021acceleratingdrugdiscoBA, yang2021efficientdrugdiscoBA, tosh2025batchie, bertin:etal:2023:recover, wang2025guide}. Existing methods for batch combination screening typically balance exploration and exploitation through uncertainty-based acquisition~\citep{lewis1995sequential, settles2009active} and retrain the predictive model after each batch. Cold-start selection of the initial batch using coverage-based methods has also been explored~\citep{de2025addressing}. Building upon these works, we propose an adaptive selection strategy that performs the exploration-exploitation tradeoff through efficacy-weighted $k$-means++ seeding in \method{}'s learned embedding space, avoiding the computational overhead of uncertainty estimation due to posterior sampling or ensemble construction~\citep{gal2016dropout, lakshminarayanan2017simple}.~\looseness=-1

% ======================================================================

% ======================================================================
\section{Experiments}
\label{sec:experiments}
% ======================================================================

\subsection{Experimental setup}
\label{sec:setup}

\paragraph{Datasets.}
We pretrain \method{} on a curated collection of 40 drug screening datasets comprising approximately 30M viability measurements across ${\sim}$3{,}700 drugs and ${\sim}$6{,}000 biological samples (full list in \cref{app:datasets}).
Pretraining uses the Adam optimizer with learning rate $10^{-5}$ for ${\sim}$400K iterations on a single NVIDIA H100 GPU; further details are in \cref{app:architecture,app:datasets}. For evaluation, we hold out four datasets not seen during pretraining.
Three are pairwise combination screens, which include observations for both drug combinations and monotherapies:
\textbf{NCI-ALMANAC}~\citep{holbeck:etal:2019:nci-almanac-combo} screens 104 drugs in ${\sim}$5{,}000 pairwise combinations across the NCI-60 cancer cell line panel;
\textbf{GDSC-SQ}~\citep{jaaks2022effective} screens 65 drugs in ${\sim}$1{,}300 combinations across 126 breast, colorectal, and pancreatic cancer cell lines;
and \textbf{BATCHIE}~\citep{tosh2025batchie} screens 207 drugs in ${\sim}$19{,}000 combinations across 16 cell lines (14 cancer plus 2 non-cancerous controls).
For faster evaluation, we use 10 cell lines from NCI-ALMANAC and GDSC-SQ.
The fourth, \textbf{PDO-Breast}~\citep{guillen2022human}, is a purely monotherapy ex vivo screen of 16 breast cancer patient-derived organoid models treated with 46 drugs (${\sim}$81K observations).
Including PDO-Breast tests generalization to the monotherapy-only setting.
For each sample in the evaluation datasets, we randomly select $n \in \{50, 100, 150, 200, 250, 300\}$ observations as few-shot context and predict viability for all remaining held-out observations.~\looseness=-1

 % and, because it uses patient-derived tissue rather than established cell lines, to the clinically relevant ex vivo regime for which no public combination datasets are currently available

% \paragraph{Evaluation protocol.}
% For each cell line, we randomly select $n \in \{50, 100, 150, 200, 250, 300\}$ observations as context and predict viability for all remaining held-out observations.
% All experiments are repeated over 3 random seeds; we report mean $\pm$ standard deviation.

\begin{figure}[htbp]
    \centering
    \includegraphics[width=\textwidth]{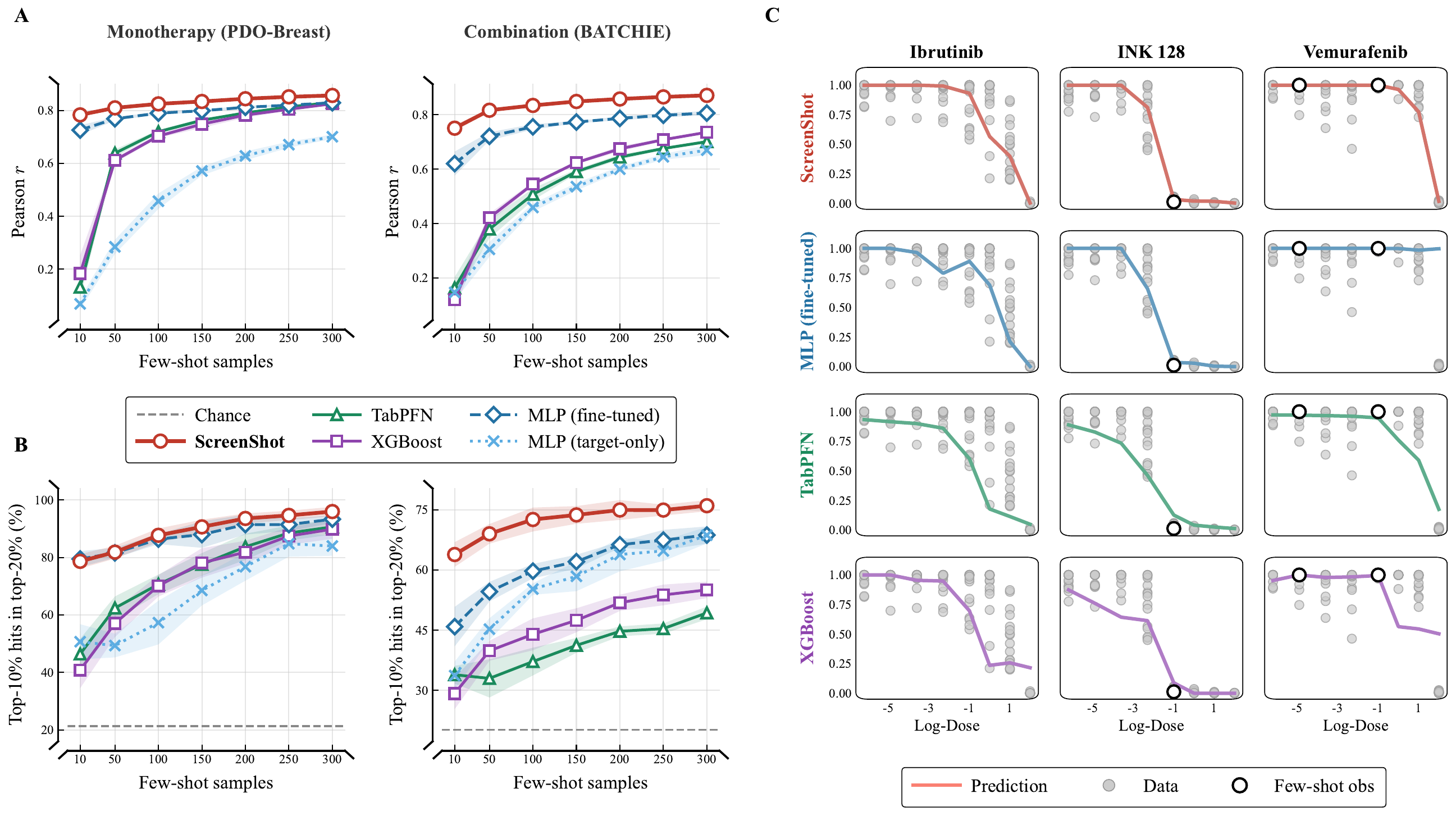}
    \caption{\textbf{Few-shot dose-response prediction on held-out datasets.}
    \textbf{(A)}~Pearson correlation as a function of the number of few-shot observations on the PDO-Breast screen (left) and BATCHIE screen (right).
    \textbf{(B)}~Top-10\% hit recall in the top-20\% predictions (PDO-Breast, left; BATCHIE, right).
    \textbf{(C)}~Example dose-response curves for three drugs on sample HCI-003 (PDO-Breast) at $n{=}100$.}
    \label{fig:fewshot}
\end{figure}

 % Gray points: held-out ground truth; white circled points: few-shot observations; colored lines: model predictions.

\paragraph{Metrics.}
We evaluate predictions along two complementary axes.
For \emph{prediction quality}, we report Pearson correlation~$r$ between predicted and true viability values.
For \emph{hit detection}, the primary practical goal in drug screening, we measure the ability to identify \emph{selectively effective} treatments: drug combinations that reduce viability in the target cancer sample more than in non-cancerous controls.
Following \citet{tosh2025batchie}, we quantify selectivity via the therapeutic index of a drug combination $\mathbf{d}$, defined as the minimum differential viability over its dose grid:
\begin{equation}
    \mathrm{TI}(\mathbf{d}, s) \;=\; \min_{\boldsymbol\delta}\;\Big[\, y_{s,(\mathbf{d},\boldsymbol\delta)} \;-\; \min_{c \in \mathcal{C}}\; y_{c,(\mathbf{d},\boldsymbol\delta)} \,\Big],
    \label{eq:ti}
\end{equation}
where $s$ is the target sample, $\mathcal{C}$ is the set of control samples, and the outer minimum selects the dose at which the target is most sensitive relative to controls.
Strongly negative $\mathrm{TI}$ indicates a selectively effective treatment.
For PDO-Breast, a monotherapy screen, we instead rank drugs by the mean differential viability across doses (i.e., the difference in area under the dose-response curve between target and control), which is a standard measure of single-drug efficacy.~\looseness=-1

For the BATCHIE dataset, $\mathcal{C}$ consists of two non-cancerous cell lines (RPE and BJ) included in the screen by design.
For NCI-ALMANAC, GDSC-SQ, and PDO-Breast, which lack designated controls, we select the cell line with the highest average viability across all treatments as a proxy control, under the assumption that non-cancerous samples are on average more resistant to drug treatment, a pattern confirmed empirically in BATCHIE, where the two non-cancerous controls rank first and third in average viability.~\looseness=-1

We define ground-truth hits as the 10\% of treatments with the most negative true $\mathrm{TI}$ per sample, and report \emph{hit recall}: the fraction of true top-10\% hits recovered among the top-20\% of treatments ranked by predicted $\mathrm{TI}$, averaged across non-control samples.
When computing predicted $\mathrm{TI}$ for evaluation, we use ground-truth viability values for the control samples.
This benefits all methods equally and reduces evaluation noise; it is also realistic, as controls are shared across targets and would be measured exhaustively in practice.
For the active learning acquisition strategy, we use model predictions for controls: using ground truth would selectively benefit adaptive strategies over random and cold-start, since only adaptive methods use the predicted delta for selection (details in \cref{app:setup_details}).~\looseness=-1

\paragraph{Baselines.}
We compare \method{} against four baselines.
\textbf{XGBoost} and \textbf{TabPFN}~\citep{hollmann2025accurate} both use 512-dimensional Morgan fingerprints~\citep{rogers2010morganfingerprint} concatenated with dose as input features; XGBoost hyperparameters are selected via 5-fold cross-validation, while TabPFN performs in-context learning.
\textbf{MLP (target-only)} is a multi-layer perceptron with learned drug and dose embeddings, trained from scratch on each target sample.
\textbf{MLP (fine-tuned)} uses the same architecture, pretrained on the same 40-dataset corpus as \method{} and fine-tuned on the target sample (see \cref{app:hyperparameters} for more details).~\looseness=-1

These baselines span the main axes of methodological variation: no pretraining (XGBoost, MLP target-only), general-purpose pretraining (TabPFN), and domain-specific pretraining with fine-tuning (MLP fine-tuned, the closest comparator to \method{}).
All baselines receive the same observations as \method{}.
A systematic comparison with existing drug screening models from the literature (\cref{app:baselines}) shows that none jointly handle combination screens, dose information, and omics-free inference, which precludes direct benchmarking.
Code, pretrained models, and an interactive dashboard are available at \url{https://github.com/tansey-lab/screenshot}.~\looseness=-1

% ---
\subsection{Few-shot prediction}
\label{sec:fewshot_results}

We first evaluate all methods with randomly selected few-shot observations.
\cref{fig:fewshot}A and B show that \method{} outperforms all baselines on both the PDO-Breast monotherapy ex vivo screen and the BATCHIE combination screen, in both Pearson correlation and hit recall, across all budgets.
The gap between \method{} and the second-best method (MLP fine-tuned) is larger on the combination screen, which is a more challenging setting due to the combinatorial treatment space.~\looseness=-1

\cref{fig:fewshot}C shows dose-response predictions for one PDO-Breast sample (HCI-003) at $n{=}100$.
The three drugs illustrate different few-shot regimes: Vemurafenib has two context observations, INK~128 has one, and Ibrutinib has none.
\method{} fits the ground truth more closely in all three cases.~\looseness=-1

\begin{table*}[t]
\centering
\scriptsize
\caption{Few-shot prediction across four held-out drug screening datasets. Values: mean$\pm$std across 3 to 5 repetitions. \textbf{Bold}: best per column within each dataset. Time: average wall-clock seconds per experiment. Bottom: average score normalized by ScreenShot (1.00 = ScreenShot level).}
\label{tab:few_shot}
\vspace{-4pt}
\newcommand{\tpm}[1]{{\tiny$\pm$#1}}
\resizebox{\textwidth}{!}{%
\setlength{\tabcolsep}{2.5pt}
\begin{tabular}{clccccc|ccccc|c}
\toprule
 &  & \multicolumn{5}{c}{Pearson $r$} & \multicolumn{5}{c}{Top-hit recall (\%)} & Time \\
\cmidrule(lr){3-7} \cmidrule(lr){8-12} \cmidrule(lr){13-13}
 &  & 10 & 50 & 100 & 200 & 300 & 10 & 50 & 100 & 200 & 300 & (s) \\
\midrule
\multirow{5}{*}{\rotatebox[origin=c]{90}{\textsc{PDO-Br.}}} & XGBoost & .18\tpm{.07} & .61\tpm{.01} & .70\tpm{.01} & .77\tpm{.01} & .82\tpm{.00} & 40.8\tpm{6.3} & 57.1\tpm{4.8} & 70.1\tpm{3.7} & 81.9\tpm{4.0} & 89.9\tpm{3.3} & 192 \\
 & TabPFN & .13\tpm{.04} & .64\tpm{.01} & .72\tpm{.01} & .79\tpm{.01} & .82\tpm{.00} & 46.4\tpm{5.3} & 62.4\tpm{4.0} & 70.7\tpm{3.4} & 83.7\tpm{4.6} & 90.7\tpm{2.7} & 127 \\
 & MLP & .07\tpm{.02} & .28\tpm{.03} & .45\tpm{.03} & .62\tpm{.02} & .69\tpm{.01} & 50.7\tpm{6.1} & 49.3\tpm{4.1} & 57.3\tpm{7.6} & 76.8\tpm{4.9} & 84.0\tpm{3.4} & 1156 \\
 & MLP-FT & .73\tpm{.02} & .77\tpm{.01} & .79\tpm{.01} & .81\tpm{.01} & .82\tpm{.00} & \bf 79.5\tpm{2.8} & 81.6\tpm{1.7} & 86.4\tpm{1.7} & 91.5\tpm{3.5} & 93.3\tpm{1.3} & 1180 \\
 & ScreenShot & \bf .78\tpm{.01} & \bf .81\tpm{.01} & \bf .82\tpm{.00} & \bf .84\tpm{.00} & \bf .85\tpm{.00} & 78.7\tpm{2.5} & \bf 81.9\tpm{2.2} & \bf 87.7\tpm{2.4} & \bf 93.6\tpm{1.7} & \bf 96.0\tpm{1.6} & \bf 8 \\[1pt]
\midrule
\multirow{5}{*}{\rotatebox[origin=c]{90}{\textsc{NCI}}} & XGBoost & .11\tpm{.03} & .25\tpm{.02} & .35\tpm{.02} & .48\tpm{.00} & .56\tpm{.01} & 24.0\tpm{1.5} & 28.5\tpm{1.7} & 33.0\tpm{4.2} & 42.6\tpm{1.5} & 50.0\tpm{0.5} & 118 \\
 & TabPFN & .16\tpm{.03} & .26\tpm{.01} & .32\tpm{.01} & .43\tpm{.01} & .50\tpm{.01} & 24.6\tpm{1.8} & 30.1\tpm{0.9} & 33.0\tpm{4.3} & 41.1\tpm{2.8} & 47.8\tpm{1.9} & 524 \\
 & MLP & .12\tpm{.01} & .18\tpm{.01} & .23\tpm{.01} & .25\tpm{.05} & .35\tpm{.02} & 24.8\tpm{2.9} & 30.0\tpm{1.7} & 32.2\tpm{1.7} & 36.4\tpm{1.8} & 40.6\tpm{1.9} & 710 \\
 & MLP-FT & .51\tpm{.01} & .57\tpm{.01} & .61\tpm{.01} & .65\tpm{.01} & .67\tpm{.01} & 34.6\tpm{0.8} & 38.2\tpm{0.8} & 40.6\tpm{1.1} & 48.5\tpm{1.5} & 49.5\tpm{2.8} & 726 \\
 & ScreenShot & \bf .58\tpm{.02} & \bf .65\tpm{.01} & \bf .68\tpm{.00} & \bf .72\tpm{.01} & \bf .74\tpm{.00} & \bf 44.5\tpm{3.9} & \bf 51.0\tpm{2.3} & \bf 54.8\tpm{1.1} & \bf 61.1\tpm{2.2} & \bf 63.0\tpm{0.6} & \bf 68 \\[1pt]
\midrule
\multirow{5}{*}{\rotatebox[origin=c]{90}{\textsc{GDSC-SQ}}} & XGBoost & .19\tpm{.04} & .47\tpm{.03} & .59\tpm{.01} & .70\tpm{.01} & .75\tpm{.01} & 26.5\tpm{2.7} & 41.9\tpm{1.8} & 51.5\tpm{4.1} & 61.0\tpm{0.5} & 66.0\tpm{2.4} & 137 \\
 & TabPFN & .28\tpm{.04} & .44\tpm{.01} & .54\tpm{.01} & .68\tpm{.01} & .75\tpm{.01} & 30.3\tpm{2.9} & 43.0\tpm{4.4} & 49.3\tpm{2.2} & 60.1\tpm{0.4} & \bf 66.3\tpm{1.2} & 655 \\
 & MLP & .25\tpm{.05} & .39\tpm{.01} & .48\tpm{.04} & .58\tpm{.02} & .66\tpm{.01} & 27.5\tpm{2.0} & 41.6\tpm{3.9} & 45.7\tpm{3.1} & 55.3\tpm{2.1} & 60.8\tpm{2.2} & 735 \\
 & MLP-FT & .51\tpm{.03} & .63\tpm{.02} & .68\tpm{.02} & .72\tpm{.01} & .76\tpm{.00} & 38.9\tpm{2.5} & 49.9\tpm{2.7} & \bf 54.8\tpm{2.2} & \bf 62.5\tpm{0.5} & 66.0\tpm{1.7} & 737 \\
 & ScreenShot & \bf .60\tpm{.02} & \bf .71\tpm{.00} & \bf .73\tpm{.01} & \bf .76\tpm{.00} & \bf .78\tpm{.01} & \bf 39.8\tpm{2.1} & \bf 51.7\tpm{1.0} & 54.4\tpm{2.9} & 58.2\tpm{0.7} & 60.6\tpm{0.6} & \bf 64 \\[1pt]
\midrule
\multirow{5}{*}{\rotatebox[origin=c]{90}{\textsc{BATCHIE}}} & XGBoost & .12\tpm{.04} & .42\tpm{.02} & .54\tpm{.01} & .67\tpm{.01} & .73\tpm{.00} & 29.2\tpm{3.9} & 39.9\tpm{2.5} & 44.0\tpm{4.0} & 51.8\tpm{2.0} & 55.0\tpm{2.0} & 196 \\
 & TabPFN & .16\tpm{.04} & .38\tpm{.03} & .50\tpm{.02} & .64\tpm{.01} & .69\tpm{.01} & 33.9\tpm{2.2} & 33.0\tpm{4.8} & 37.2\tpm{3.5} & 44.7\tpm{1.2} & 49.3\tpm{1.6} & 432 \\
 & MLP & .15\tpm{.03} & .30\tpm{.03} & .45\tpm{.01} & .59\tpm{.01} & .66\tpm{.02} & 33.7\tpm{3.5} & 45.2\tpm{2.7} & 55.3\tpm{1.4} & 63.9\tpm{4.2} & 68.7\tpm{2.1} & 1152 \\
 & MLP-FT & .62\tpm{.04} & .72\tpm{.01} & .75\tpm{.01} & .78\tpm{.01} & .80\tpm{.01} & 45.9\tpm{4.9} & 54.6\tpm{2.4} & 59.7\tpm{1.8} & 66.3\tpm{0.8} & 68.7\tpm{2.1} & 1142 \\
 & ScreenShot & \bf .75\tpm{.01} & \bf .82\tpm{.00} & \bf .83\tpm{.01} & \bf .86\tpm{.00} & \bf .87\tpm{.00} & \bf 63.9\tpm{3.1} & \bf 69.0\tpm{2.4} & \bf 72.6\tpm{3.0} & \bf 75.0\tpm{2.5} & \bf 76.0\tpm{1.4} & \bf 22 \\[1pt]
\midrule
\multirow{5}{*}{\rotatebox[origin=c]{90}{\textsc{Avg}}} & XGBoost & 0.23 & 0.58 & 0.71 & 0.82 & 0.88 & 0.54 & 0.66 & 0.74 & 0.83 & 0.89 & 9.48 \\
 & TabPFN & 0.28 & 0.57 & 0.67 & 0.79 & 0.85 & 0.61 & 0.67 & 0.71 & 0.80 & 0.86 & 13.59 \\
 & MLP & 0.23 & 0.39 & 0.52 & 0.64 & 0.72 & 0.60 & 0.66 & 0.71 & 0.80 & 0.86 & 56.23 \\
 & MLP-FT & 0.87 & 0.90 & 0.92 & 0.94 & 0.95 & 0.87 & 0.88 & 0.89 & 0.93 & 0.94 & 56.98 \\
 & ScreenShot & \bf 1.00 & \bf 1.00 & \bf 1.00 & \bf 1.00 & \bf 1.00 & \bf 1.00 & \bf 1.00 & \bf 1.00 & \bf 1.00 & \bf 1.00 & \bf 1.00 \\
\bottomrule
\end{tabular}}
\end{table*}

\cref{tab:few_shot} extends the comparison to all four held-out datasets.
\method{} achieves the highest Pearson correlation on every dataset and budget, and the highest hit recall on 16 out of 20 dataset/budget combinations.
The advantage is largest at low budgets, where pretraining matters most: at $n{=}10$ on BATCHIE, \method{} reaches 63.9\% hit recall compared to 45.9\% for the next-best method, and on NCI-ALMANAC the gap over MLP fine-tuned exceeds 13 percentage points at $n{=}300$.
\method{} is also 5--50$\times$ faster than all baselines.~\looseness=-1

% , requiring only a single forward pass (8--68\,s) compared to per-sample training (2--20\,min).

\FloatBarrier
% ---
\subsection{Active learning for hit detection}
\label{sec:active_results}

We now compare different strategies for selecting the few-shot observations, all using \method{} as the underlying predictor.
We consider four strategies: \emph{random} selection (the baseline used in \cref{sec:fewshot_results}); \emph{cold-start} selection via $k$-medoids on drug-dose embeddings; and two \emph{adaptive} strategies that allocate 33\% of the budget to a cold-start batch and the remainder to delta-guided adaptive rounds (\cref{alg:active_learning}), with either one round (batch size = 67\% of budget) or two rounds (two batches of 33\%).~\looseness=-1

\begin{figure}[htbp]
    \centering
    \includegraphics[width=\textwidth]{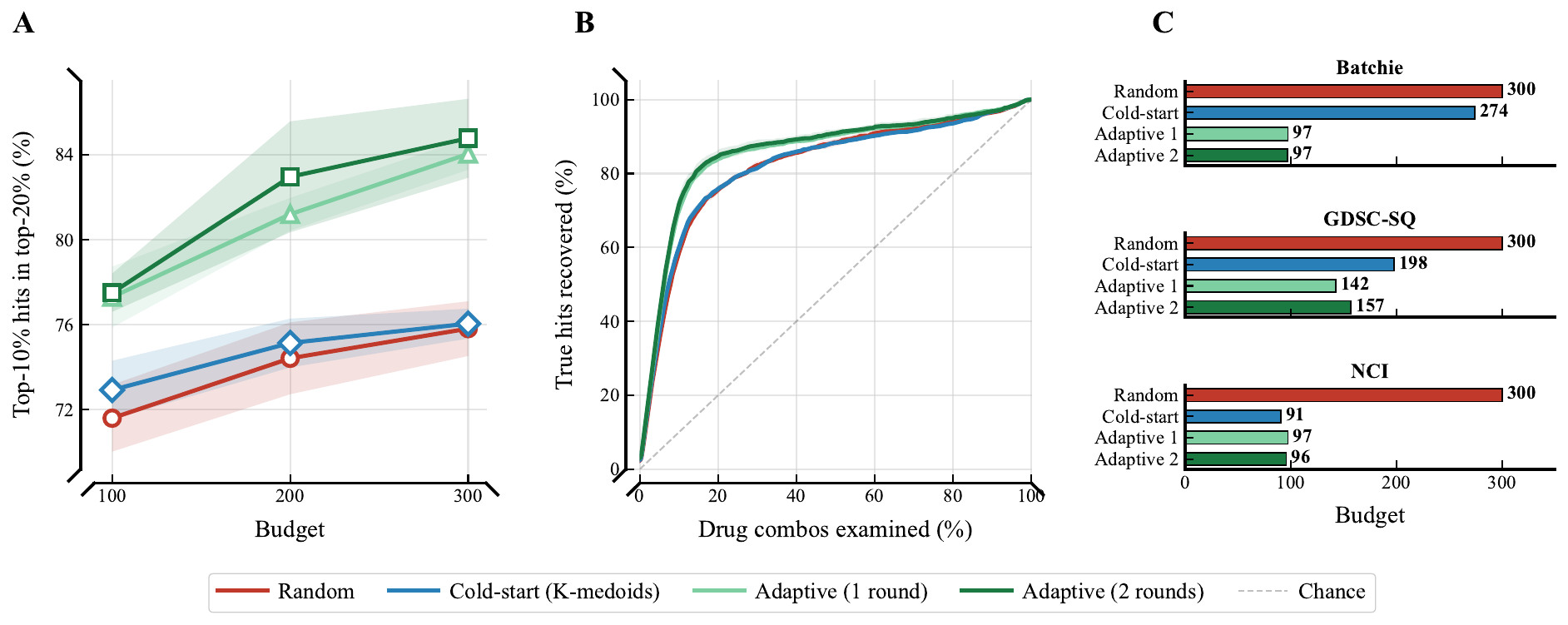}
    \caption{\textbf{Active learning for hit identification.}
    \textbf{(A)}~Hit recall (top-10\% hits in top-20\% predictions) as a function of total budget on BATCHIE.
    \textbf{(B)}~Cumulative hit recall curves at budget~300 on BATCHIE, showing that the ranking is robust to the choice of prediction set size.
    \textbf{(C)}~Budget required to match random selection at budget~300, across three datasets.}
    \label{fig:active_learning}
\end{figure}

\cref{fig:active_learning}A shows that the adaptive approach yields a significant improvement on BATCHIE: at budget~300, adaptive selection reaches ${\sim}$85\% hit recall compared to ${\sim}$76\% for both random and cold-start selection, a 9~percentage-point gain.
Cold-start selection alone does not significantly improve over random on this dataset.
\cref{fig:active_learning}B shows cumulative recall curves at budget~300: the adaptive variants recover a larger fraction of true hits at every prediction set size, confirming that the improvement is robust to the choice of top-$k$ threshold.

\cref{fig:active_learning}C extends the analysis to three datasets via a budget equivalence comparison: for each method, we report the budget needed to match the hit recall of random selection at budget~300 (obtained by linear interpolation between evaluated budgets).
Adaptive strategies require 2--3$\times$ less budget than random to reach the same recall.
Cold-start selection helps on some datasets (notably NCI).
The number of adaptive rounds has a modest effect; the dominant factors are the total budget and whether adaptive selection is used at all.
This is encouraging for clinical applications, where sample availability naturally limits the number of experimental rounds.~\looseness=-1

% ======================================================================
\section{Discussion and limitations}
\label{sec:discussion}
% ======================================================================

\paragraph{Drug coverage.}
\method{} is designed to assist in large-scale pharmacological screens for compounds that have experimental data in the literature.
The pretrained library covers ${\sim}$3{,}700 unique drugs resolved through PubChem, providing at least 89\% coverage on all four evaluation datasets (\cref{app:drug_coverage}).
For novel compounds, the model uses pretrained UNK tokens that allow it to leverage few-shot context observations to infer the drug's behavior; we evaluate the effect of UNK tokens compared to alternatives in \cref{app:drug_coverage} and show that they are beneficial.
Compared to methods relying on Morgan fingerprints as drug representation in our benchmark, \method{} consistently performs better.
As the corpus of public drug screening data grows, the pretrained library will naturally expand.~\looseness=-1

\paragraph{Architecture.}
An ablation study (\cref{app:arch_ablation}) confirms that each level of the hierarchy improves prediction quality and that the hierarchical design is faster than flat attention.
Removing the Level~3 self-attention reduces inference time with only a modest drop in hit recall, suggesting that future work could explore lighter alternatives to the sample-level self-attention while preserving accuracy.~\looseness=-1

\paragraph{Accuracy vs.\ hit detection.}
The active learning ablation (\cref{app:ablation}) reveals that global prediction accuracy and hit detection ability are not directly linked.
Using MC dropout uncertainty as selection weights instead of predicted delta yields the best MAE and Pearson correlation but 6 percentage points lower hit recall on BATCHIE (78.6\% vs.\ 84.8\% at budget 300).
This suggests that uncertainty-guided selection targets regions of model ignorance rather than regions of differential drug sensitivity.
When the goal is to identify top hits, weighting by the objective of interest (therapeutic index) is more effective than reducing global uncertainty.~\looseness=-1

% \paragraph{Diversity in selection.}
% Pure exploitation strategies that greedily select the top-$k$ treatments by predicted delta or uncertainty perform substantially worse than their diversity-aware counterparts.
% Clustering methods like $k$-medoids and $k$-means++ provide essential diversity and avoid redundant observations that fail to improve the model globally.
% \cref{app:ablation} provides additional ablation results.

% ======================================================================
% \begin{ack}
% \todo{Add acknowledgments and funding disclosure.}
% \end{ack}
% ======================================================================

% ======================================================================
\section*{Acknowledgements}
\label{sec:intro}
% ======================================================================
WT is supported by the NIH/NCI (R37 CA271186, U54 CA274492, P30 CA008748), Break Through Cancer, the Fund for Innovation in Cancer Informatics, the Cancer AI Alliance, the Tow Center for Developmental Oncology, and the Maurice Campbell Initiative at Memorial Sloan Kettering Cancer Center.

% \clearpage
\bibliographystyle{unsrtnat}
% \bibliography{references}

% ======================================================================
\newpage
\appendix
% ======================================================================
% Appendix
% ======================================================================

\section{Model architecture details}
\label{app:architecture}

\begin{figure}[htbp]
\centering
\resizebox{0.85\textwidth}{!}{%
\begin{tikzpicture}[
    >=Stealth,
    node distance=0.5cm,
    box/.style={draw, rounded corners=2pt, minimum width=3.6cm, minimum height=0.7cm,
                font=\small, align=center},
    add/.style={circle, draw, inner sep=1.5pt, font=\scriptsize},
    shared/.style={densely dashed, gray, thick},
]

% Column centers
\def\colL{0}      % Query column (left)
\def\colR{5.5}    % Context column (right)
\def\colLvl{10.2} % Level labels (right margin)

% ===================== Headers =====================
\node[font=\small\bfseries] at (\colL, 0) (hdr_q) {Query};
\node[font=\small\bfseries] at (\colR, 0) (hdr_c) {Context};

% ===================== LEVEL 1 =====================

% --- Drug Embedding ---
\node[box] at (\colL, -1.0) (q_drug) {Drug Embedding\\[-2pt]{\scriptsize $(q) \to (q, 256)$}};
\node[box] at (\colR, -1.0) (c_drug) {Drug Embedding\\[-2pt]{\scriptsize $(p, m, q) \to (p, m, q, 256)$}};
\draw[shared] (q_drug) -- (c_drug);

% --- Dose Embedding (mirrored on both sides) ---
\node[box, minimum width=2.4cm] at (\colL-2.7, -2.2) (q_dose) {Dose Emb.\\[-2pt]{\scriptsize $(q) \to (q, 256)$}};
\node[box, minimum width=2.4cm] at (\colLvl-2.0, -2.2) (c_dose) {Dose Emb.\\[-2pt]{\scriptsize $(p, m, q) \to (p, m, q, 256)$}};

% --- Sum with dose ---
\node[add] at (\colL, -2.2) (q_add1) {$+$};
\node[add] at (\colR, -2.2) (c_add1) {$+$};
\draw[->, thick] (q_drug) -- (q_add1);
\draw[->, thick] (c_drug) -- (c_add1);
\draw[->, thick] (q_dose.east) -- (q_add1);
\draw[->, thick] (c_dose.west) -- (c_add1);

% --- LayerNorm ---
\node[box] at (\colL, -3.9) (q_ln) {LayerNorm\\[-2pt]{\scriptsize $(q, 256)$}};
\node[box] at (\colR, -3.9) (c_ln) {LayerNorm\\[-2pt]{\scriptsize $(p, m, q, 256)$}};
\draw[->, thick] (q_add1) -- (q_ln);
\draw[->, thick] (c_add1) -- (c_ln);
\draw[shared] (q_ln) -- (c_ln);

% --- Self-Attention x5 ---
\node[box] at (\colL, -5.1) (q_sa1) {Self-Attention $\times$ 5\\[-2pt]{\scriptsize $(q, 256) \to (q, 256)$}};
\node[box] at (\colR, -5.1) (c_sa1) {Self-Attention $\times$ 5\\[-2pt]{\scriptsize $(p, m, q, 256) \to (p, m, q, 256)$}};
\draw[->, thick] (q_ln) -- (q_sa1);
\draw[->, thick] (c_ln) -- (c_sa1);
\draw[shared] (q_sa1) -- (c_sa1);

% --- Masked Average ---
\node[box] at (\colL, -6.3) (q_avg) {Masked Average\\[-2pt]{\scriptsize $(q, 256) \to (256)$}};
\node[box] at (\colR, -6.3) (c_avg) {Masked Average\\[-2pt]{\scriptsize $(p, m, q, 256) \to (p, m, 256)$}};
\draw[->, thick] (q_sa1) -- (q_avg);
\draw[->, thick] (c_sa1) -- (c_avg);

% ===================== LEVEL 2 =====================

% --- Viability injection on context (from right) ---
\node[box, minimum width=2.4cm] at (\colLvl-2.0, -7.5) (vib_emb) {Viability Emb.\\[-2pt]{\scriptsize $(p, m) \to (p, m, 256)$}};
\node[add] at (\colR, -7.5) (c_add2) {$+$};
\draw[->, thick] (c_avg) -- (c_add2);
\draw[->, thick] (vib_emb.west) -- (c_add2);

% --- Cross-Attention ---
\node[box, minimum width=9.3cm] at ({(\colL+\colR)/2}, -8.8) (ca) {Cross-Attention $\times$ 5\\[-2pt]{\scriptsize Q: $(p, 1, 256)$, \; K,V: $(p, m, 256)$ \; $\to (p, 1, 256)$}};
\draw[->, thick] (q_avg) -- (q_avg |- ca.north) node[midway, left, font=\scriptsize\itshape] {Q};
\draw[->, thick] (c_add2) -- (c_add2 |- ca.north) node[midway, right, font=\scriptsize\itshape] {K, V};

% ===================== LEVEL 3 =====================

\node[box, minimum width=5cm] at ({(\colL+\colR)/2}, -10.2) (sa3) {Self-Attention $\times$ 5\\[-2pt]{\scriptsize $(p, 256) \to (p, 256)$}};
\draw[->, thick] (ca) -- (sa3);

\node[box, minimum width=5cm] at ({(\colL+\colR)/2}, -11.3) (cls) {Pooling\\[-2pt]{\scriptsize $(p, 256) \to (1, 256)$}};
\draw[->, thick] (sa3) -- (cls);

% ===================== HEAD =====================

\node[box, minimum width=5cm] at ({(\colL+\colR)/2}, -12.4) (head) {Linear + Sigmoid\\[-2pt]{\scriptsize $(256) \to \hat{y} \in [0,1]$}};
\draw[->, thick] (cls) -- (head);

% Lock bounding box to original diagram extent (before annotations)
\useasboundingbox (\colL-2.7-1.5, -12.9) rectangle (\colLvl, 0.5);

% ===================== ANNOTATIONS =====================
% Encoder labels (left side, vertically aligned)
% Encoder labels (left, right-aligned so ends line up, close to boxes)
\node[font=\small\bfseries, text=gray!50, anchor=east] at (\colL-2.3, -5.1) {DCE};
\node[font=\small\bfseries, text=gray!50, anchor=east] at (\colL-2.3, -8.8) {RE};
\node[font=\small\bfseries, text=gray!50, anchor=east] at (\colL-2.3, -10.2) {SE};

% Level labels (right, left-aligned, close to boxes)
% Level 0 removed
\node[font=\small\bfseries, text=gray!50, anchor=west] at (\colLvl-1.5, -5.1) {Level 1};
\node[font=\small\bfseries, text=gray!50, anchor=west] at (\colLvl-1.5, -8.8) {Level 2};
\node[font=\small\bfseries, text=gray!50, anchor=west] at (\colLvl-1.5, -10.2) {Level 3};

\end{tikzpicture}
}%
\caption{Layer-by-layer architecture of \method{} (7.0M parameters). Query and context follow identical pipelines through Level~1 (dashed lines = shared weights). DCE = Drug Combination Encoder, RE = Response Encoder, SE = Sample Encoder. $q$: drugs per combination, $p$: unique perturbations, $m$: observations per perturbation. Dose and viability use trainable Fourier features. Drug embedding vocabulary: 3,779 entries (3,679 drugs + 100 UNK).}
\label{fig:arch_diagram}
\end{figure}
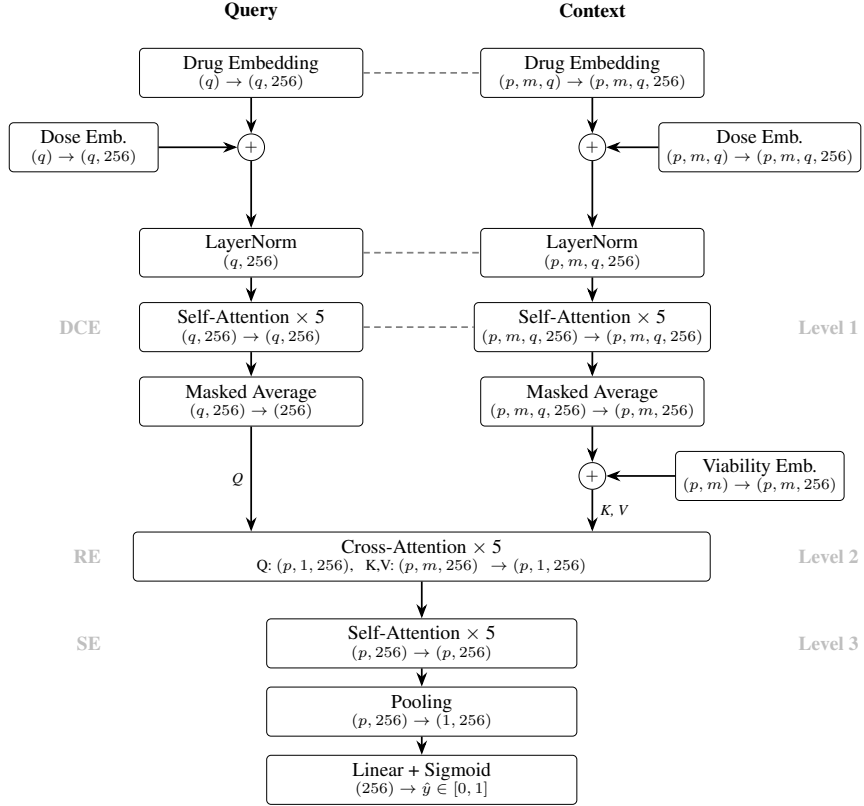

\cref{fig:arch_diagram} provides a layer-by-layer view of the \method{} architecture (${\sim}$7M parameters).
All three encoders (Drug Combination Encoder, DCE; Response Encoder, RE; Sample Encoder, SE) share the same transformer block configuration: 5 layers, 8 attention heads (head dim 32), hidden dim 256, GeLU activations, residual connections and layer normalization.
Given $n$ context observations for a sample, we group them by perturbation: observations from the same drug combination are placed together.
This yields $p$ unique perturbations, each with up to $m$ dose-level observations, where each perturbation involves up to $q$ drugs.
The context is stored as a padded tensor of shape $(p, m, q)$; $p$, $m$, and $q$ are the maximum values across the batch, and padding masks are applied at every level.
In particular, $q$ is padded because datasets often contain both monotherapy ($q{=}1$) and combination ($q{=}2$ or $3$) observations.
A learnable CLS token is appended to the perturbation sequence before the Sample Encoder; its output after self-attention is used as the pooled sample representation.

\paragraph{Fourier encoding.}
Continuous scalars (doses and viabilities) are encoded using learned Fourier features:
\begin{equation}
    \phi(x) = W \bigl[\sin(x \cdot \boldsymbol{\omega}),\; \cos(x \cdot \boldsymbol{\omega})\bigr] + \mathbf{b},
\end{equation}
where $\boldsymbol{\omega}\in\mathbb R^{F}$, $W\in\mathbb R^{D\times 2F}$, and $\mathbf b\in\mathbb R^{D}$ are learnable parameters (with independent parameters for doses and for viabilities). We use $F{=}64$ Fourier frequencies.

% ======================================================================
\FloatBarrier
\section{Pretraining}
\label{app:datasets}

\paragraph{Pretraining corpus.}
\cref{tab:pretraining_datasets} lists all datasets in the pretraining and evaluation corpus.
The ex vivo datasets were obtained on request from \citet{pichotta2026pan}, and we added publicly available cell line datasets for a total of 40 studies spanning both monotherapy and combination screens across diverse cancer types.

\begin{table}[htbp]
    \caption{Pretraining corpus: 40 drug screening datasets. $^\dagger$Held out for evaluation. CL = cell line, EV = ex vivo (patient-derived). $^*$Number of unique consensus drug IDs after PubChem resolution.}
    \label{tab:pretraining_datasets}
    \centering
    \scriptsize
    \begin{tabular}{llrrrll}
        \toprule
        \textbf{Dataset} & \textbf{Reference} & \textbf{Samples} & \textbf{Drugs} & \textbf{Obs.} & \textbf{Model} & \textbf{Type} \\
        \midrule
        Beat AML & \citet{bottomly:etal:2022:beat-aml-pdc} & 631 & 166 & 491K & EV & Mono \\
        Nair et al. & \citet{nair:etal:2023:nsclc-cl} & 81 & 244 & 3.6M & CL & Combi \\
        Betge et al. & \citet{betge:etal:2022:crc_pdo} & 14 & 520 & 26K & EV & Mono \\
        Martins et al. & \citet{martins:etal:2022:hgsoc-pdc} & 25 & 12 & 6K & EV & Mono \\
        Bruna et al. & \citet{bruna:etal:2016:breast-pdx-pdxc} & 21 & 105 & 42K & EV & Combi \\
        CCLE & \citet{barretina:etal:2012:ccle} & 504 & 24 & 93K & CL & Mono \\
        Driehuis et al. & \citet{driehuis:etal:2019:panc-pdo} & 24 & 78 & 35K & EV & Combi \\
        CTRP1 & \citet{basu:etal:2013:ctrp1} & 242 & 203 & 257K & CL & Mono \\
        CTRP2 & \citet{seashore-ludlow:etal:2015:ctrp2} & 887 & 545 & 12.4M & CL & Mono \\
        Mayoh et al. & \citet{mayoh:etal:2023:pediatric} & 151 & 144 & 35K & EV & Mono \\
        Friedman et al.\,'15 & \citet{friedman:etal:2015:melanoma-pdc-comb} & 37 & 108 & 419K & EV & Combi \\
        Friedman et al.\,'17 & \citet{friedman:etal:2017:melanoma-pdc-pdx} & 20 & 76 & 105K & EV & Combi \\
        GDSC1 & \citet{yang:etal:2013:gdsc} & 988 & 304 & 2.6M & CL & Mono \\
        GDSC2 & \citet{yang:etal:2013:gdsc} & 810 & 169 & 1.4M & CL & Mono \\
        Murumägi et al. & \citet{murumagi:etal:2022:ovarian-pdc} & 13 & 641 & 45K & EV & Mono \\
        Sa et al. & \citet{sa:etal:2019:gyn-pdc} & 45 & 97 & 42K & EV & Mono \\
        Yan et al. & \citet{yan:etal:2018:gastric-pdo} & 9 & 37 & 7K & EV & Mono \\
        SMarTrial & \citet{smartrial} & 180 & 108 & 66K & EV & Combi \\
        Lau et al. & \citet{lau:etal:2021:pediatric-pdx-pdc} & 17 & 194 & 21K & EV & Mono \\
        Ice et al. & \citet{ice:etal:2019:melanoma-pdx-pdxc} & 8 & 38 & 5K & EV & Mono \\
        Merck & \citet{oneil:etal:2016:merck-cl-combo} & 39 & 38 & 1.5M & CL & Combi \\
        Lee et al.\,'18 & \citet{lee:etal:2018:pancan-pdc} & 284 & 67 & 234K & EV & Mono \\
        Johansson et al. & \citet{johansson:etal:2020:gbm-pdc} & 86 & 265 & 202K & EV & Mono \\
        Ronteix et al. & \citet{ronteix2025cohort} & 158 & 29 & 35K & EV & Mono \\
        Powell et al. & \citet{powell:etal:2020:tnbc-pdxc} & 16 & 559 & 118K & EV & Mono \\
        Malani et al. & \citet{malani:etal:2021:aml-precision-onc} & 181 & 539 & 369K & EV & Mono \\
        Toshimitsu et al. & \citet{toshimitsu:etal:2022:crc-pdo} & 26 & 56 & 31K & EV & Mono \\
        Hirt et al. & \citet{hirt:etal:2022:pdac-pdo} & 18 & 25 & 3K & EV & Mono \\
        Li et al. & \citet{li:etal:2019:liver:pdo} & 27 & 129 & 3K & EV & Mono \\
        Lee et al.\,'18b & \citet{lee:etal:2018:bladder-pdo} & 11 & 21 & 5K & EV & Mono \\
        Al Shihabi et al. & \citet{soragni:pdo_sarcoma_2024} & 109 & 530 & 21K & EV & Combi \\
        Gu et al. & \citet{gu:etal:2022:head-neck-pdc} & 13 & 2,245 & 34K & EV & Mono \\
        Polley et al. & \citet{polley:etal:2016:sclc-pdc} & 70 & 523 & 389K & CL & Mono \\
        Tiriac et al. & \citet{tiriac:etal:2018:pancreas-pdo} & 68 & 5 & 14K & EV & Mono \\
        Peterziel et al. & \citet{peterziel:etal:2022:peditaric-pdc} & 65 & 79 & 21K & EV & Mono \\
        Pemovska et al. & \citet{pemovska:etal:2013:refractory-aml-pdc} & 35 & 222 & 31K & EV & Mono \\
        \midrule
        BATCHIE$^\dagger$ & \citet{tosh2025batchie} & 16 & 207 & 148K & CL & Combi \\
        GDSC-SQ$^\dagger$ & \citet{jaaks2022effective} & 126 & 65 & 3.7M & CL & Combi \\
        NCI-ALMANAC$^\dagger$ & \citet{holbeck:etal:2019:nci-almanac-combo} & 60 & 102 & 3.4M & CL & Combi \\
        PDO-Breast$^\dagger$ & \citet{guillen2022human} & 16 & 46 & 81K & EV & Mono \\
        \midrule
        \textbf{Total} & & \textbf{6,135} & \textbf{3,679$^*$} & \textbf{32.1M} & & \\
        \bottomrule
    \end{tabular}
\end{table}

\paragraph{Data preprocessing.}
All viability values are clipped to $[0, 1]$.
Dose concentrations are expressed in micromolar ($\mu$M) and converted to $\log_{10}(\mu\text{M})$ scale, clipped to $[-6, 4]$ (corresponding to $10^{-6}$--$10^{4}\;\mu$M), and linearly rescaled to $[0, 1]$.
Drug names across datasets are resolved to consensus identifiers via PubChem lookup (matching by compound ID, substance ID, parent compound, or canonical SMILES).
To remove noisy observations, we fit a monotone decreasing isotonic regression to each (sample, drug combination) dose-response curve and discard curves with MAE above 0.2, which removes approximately 5\% of the data.

\paragraph{Leave-one-out evaluation.}
For each evaluation dataset, we train a separate foundation model on all remaining datasets.
The BATCHIE and PDO-Breast evaluations use the same pretrained model (trained on all datasets excluding BATCHIE and PDO-Breast).
For GDSC-SQ and NCI-ALMANAC, we train on all datasets except the evaluation set itself.
All models use identical hyperparameters (\cref{app:hyperparameters}).

\paragraph{Data sampling.}
At each training step, one dataset is sampled uniformly, and a batch is constructed by randomly selecting 1--16 biological samples, up to 100 unique drug combinations per sample, and up to 10 observations per combination.
A query set of 32 observations per sample is drawn separately for supervision.
Each dataset contributes 100 batches per epoch, yielding ${\sim}$4{,}000 batches per epoch and ${\sim}$400K iterations over 100 epochs.

\paragraph{UNK token training.}
To prepare the model for unknown drugs at inference time, a random fraction $\alpha \sim \text{Uniform}[0, 0.2]$ of unique drug identities per batch are replaced with one of the 100 trainable UNK tokens during pretraining.
This teaches the model to produce reasonable predictions even when a drug has no specific embedding, by relying on the few-shot context observations to infer the drug's behavior.

\begin{figure}[htbp]
    \centering
    \includegraphics[width=0.7\columnwidth]{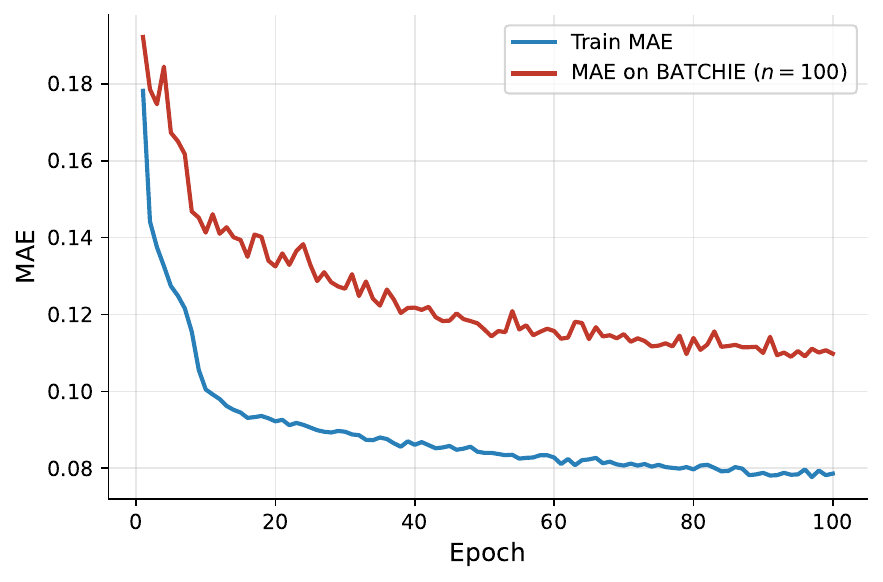}
    \caption{Foundation model training curve (BATCHIE). Train MAE and held-out MAE on BATCHIE over 100 epochs, where the held-out evaluation uses $n{=}100$ few-shot context observations per sample, uniformly sampled.}
    \label{fig:training_curve}
\end{figure}

\paragraph{Training curve.}
\cref{fig:training_curve} shows the training curve for the BATCHIE foundation model.
Both train and held-out MAE decrease steadily throughout training with no sign of overfitting.

% ======================================================================
\FloatBarrier
\section{Drug coverage}
\label{app:drug_coverage}

\cref{tab:drug_coverage} reports the overlap between each pretrained drug library and the drugs in the corresponding evaluation dataset.
Each model's library contains ${\sim}$3{,}650--3{,}680 unique consensus drug IDs, resolved through PubChem to handle name aliases across studies.
Drug names are mapped to PubChem compound IDs (CIDs) and substance IDs (SIDs); drugs sharing the same CID, SID, parent compound, or canonical SMILES are assigned a single consensus ID.
Coverage ranges from 89.4\% (BATCHIE) to 100\% (GDSC-SQ), with most evaluation drugs already present in the pretrained vocabulary.
The uncovered drugs are investigational compounds that appear in few studies (e.g., CYC065, LY3295668) and are therefore absent from the pretraining corpus.
These drugs are assigned dedicated UNK embeddings.

\begin{table}[h]
\caption{\textbf{Drug coverage.} For each evaluation dataset, the number of unique drugs, the pretrained library size (consensus drug IDs), and the number of drugs resolved to a known embedding vs.\ assigned a UNK token.}
\label{tab:drug_coverage}
\centering
\small
\begin{tabular}{lrrrr}
\toprule
\textbf{Dataset} & \textbf{Library size} & \textbf{Eval drugs} & \textbf{Known (\%)} & \textbf{UNK} \\
\midrule
BATCHIE & 3,654 & 207 & 185 (89\%) & 22 \\
GDSC-SQ & 3,679 & 65 & 65 (100\%) & 0 \\
NCI-ALMANAC & 3,675 & 102 & 98 (96\%) & 4 \\
PDO-Breast & 3,654 & 46 & 44 (96\%) & 2 \\
\bottomrule
\end{tabular}
\end{table}

\begin{figure}[htbp]
    \centering
    \includegraphics[width=\textwidth]{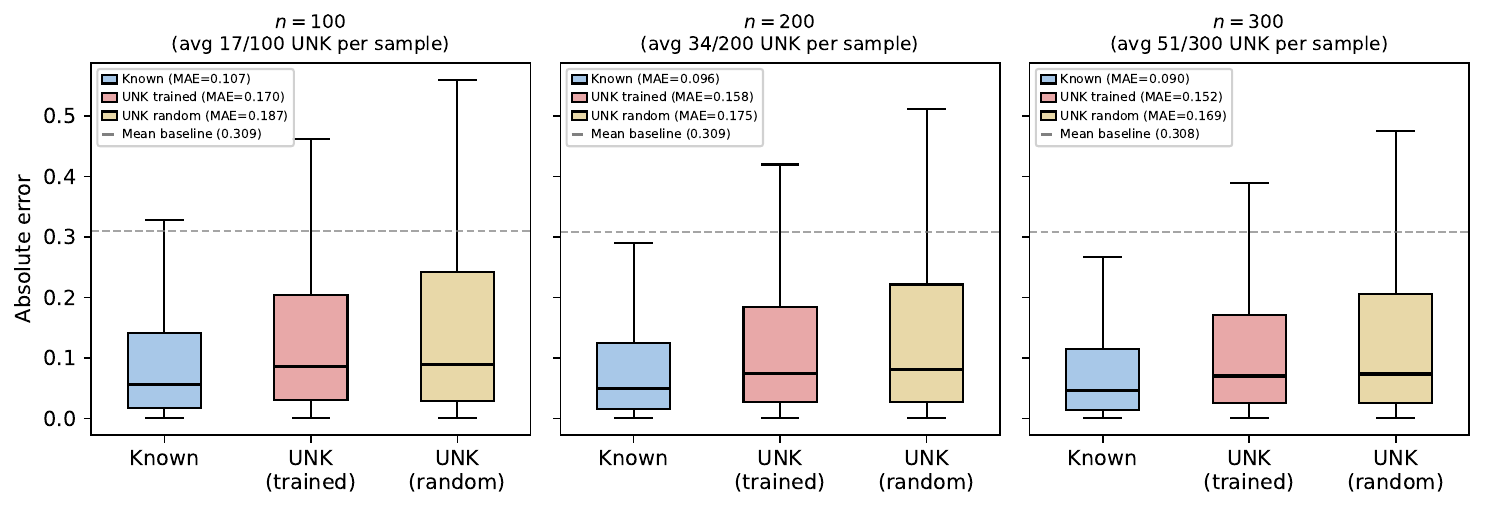}
    \caption{\textbf{Prediction error for known vs.\ unknown drugs on BATCHIE.}
    Each panel corresponds to a different few-shot budget ($n \in \{100, 200, 300\}$).
    Three conditions: treatments with known drugs (blue), treatments with UNK drugs using pretrained UNK embeddings (pink), and treatments with UNK drugs using random known drug tokens (yellow).
    Dashed line: MAE of predicting the mean few-shot viability.}
    \label{fig:unk_vs_known}
\end{figure}

\cref{fig:unk_vs_known} compares prediction error on treatments involving known drugs versus unknown (UNK) drugs on the BATCHIE dataset.
UNK drugs account for approximately 17\% of few-shot observations across all budgets.
UNK predictions remain well below the mean-prediction baseline (0.308 MAE), though naturally less accurate than known drugs (0.152 vs.\ 0.090 at budget~300).
To assess whether using a pretrained UNK token is useful, we compare it against assigning a random known drug token to each unknown drug.
For each budget (100, 200, 300), we run \method{} inference on BATCHIE with randomly selected few-shot context and compute the absolute error on all remaining observations.
We repeat with 5 random seeds, each with a different random UNK-to-known-drug assignment, and average the error across seeds.
Random tokens perform worse than pretrained UNK embeddings (+0.017 MAE at budget~300), confirming that an incorrect drug identity is more harmful than a non-specific placeholder, because a wrong identity may actively mislead the model's dose-response expectations.

% ======================================================================
\FloatBarrier
\section{Experimental setup details}
\label{app:setup_details}

\subsection{Samples}
\label{app:samples}
\cref{tab:eval_samples} lists the biological samples used for evaluation in each dataset.
BATCHIE and PDO-Breast use all available samples.
For NCI-ALMANAC and GDSC-SQ, we randomly select 10 cell lines for faster evaluation.

\begin{table}[h]
    \caption{Evaluation samples and control samples per dataset.}
    \label{tab:eval_samples}
    \centering
    \small
    \begin{tabular}{lrll}
        \toprule
        \textbf{Dataset} & \textbf{$N$} & \textbf{Control(s)} & \textbf{Sample IDs} \\
        \midrule
        BATCHIE & 16 & RPE, BJ & A673, Kelly, MDA-MB-231, MG-63, MSKEWS-33838, \\
        & & & MSKEWS-66647, MSKOST-11890, MSKRMS-12808, \\
        & & & SAOS-2, SJSA-1, SKNEP, TC-71, U2OS, Wit49 \\
        \midrule
        PDO-Breast & 16 & HCI-011 & HCI-001, HCI-002, HCI-003, HCI-005, HCI-008, HCI-010, \\
        & & & HCI-012, HCI-015, HCI-016, HCI-017, \\
        & & & HCI-019, HCI-023, HCI-024, HCI-025, HCI-027 \\
        \midrule
        NCI-ALMANAC & 10 & UACC-257 & 786-0, CAKI-1, HCT-15, HL-60(TB), NCI-H522, \\
        & & & OVCAR-4, SK-MEL-2, SK-OV-3, T-47D \\
        \midrule
        GDSC-SQ & 10 & MDA-MB-157 & CAL-148, DU-4475, HCC1395, HCC1599, HCC2218, \\
        & & & HCC70, HDQ-P1, MCF7, T47D \\
        \bottomrule
    \end{tabular}
\end{table}

\paragraph{Control samples.}
For BATCHIE, two non-cancerous cell lines (RPE and BJ) serve as controls; the control baseline is $\min(\text{RPE}, \text{BJ})$ per treatment.
For NCI-ALMANAC, GDSC-SQ, and PDO-Breast, which lack designated controls, we use the cell line with the highest average viability across all treatments as a proxy control: UACC-257, MDA-MB-157, and HCI-011 respectively.
This selection is based on the assumption that less drug-sensitive samples approximate non-cancerous controls.

\subsection{Few-shot protocol}
\label{app:fewshot_protocol}
For each sample, we randomly select $n \in \{10, 50, 100, 150, 200, 250, 300\}$ observations as context and predict viability for all remaining observations.
Few-shot experiments use 5 random seeds on BATCHIE and PDO-Breast, and 3 seeds on GDSC-SQ and NCI-ALMANAC.

\subsection{Active learning protocol}
\label{app:al_protocol}
The adaptive strategy (\cref{alg:active_learning}) proceeds in multiple rounds.
Active learning experiments use 5 seeds on all three datasets.
Observations are selected uniformly at random without replacement within each sample.
A fraction of the total budget (33\% in the main experiments) is allocated to a cold-start batch; the remainder is split equally across 1 or 2 adaptive rounds.

\paragraph{Cold-start selection.}
The cold-start batch is selected by $k$-medoids clustering on the Level~1 drug-dose embeddings (output of the DCE, before any viability is observed).
$k$-medoids selects $k$ representatives from the candidate set by solving
\begin{equation}
    \min_{S \subset \{1,\ldots,N\},\; |S|=k} \sum_{i=1}^{N} w_i \min_{j \in S} \| \mathbf{h}_i - \mathbf{h}_j \|^2,
    \label{eq:kmedoids}
\end{equation}
where $\mathbf{h}_i$ is the embedding of treatment $i$ and $w_i$ is a per-treatment weight (uniform for cold start).
We use the OneBatchPAM implementation~\citep{deMathelin2025onebatchpam}.
For control samples, we always use $k$-medoids with uniform weights regardless of the strategy used for target samples, since the goal for controls is to maximize coverage rather than target specific hits.

\paragraph{Adaptive selection.}
In adaptive rounds, we use $k$-means++ seeding on the Level~3 context-conditioned embeddings (output of the SE after observing the current context).
$k$-means++ sequentially samples centers with probability proportional to the weighted squared distance to the nearest existing center:
\begin{equation}
    P(i) \propto w_i \min_{j \in S} \| \mathbf{h}_i - \mathbf{h}_j \|^2,
    \label{eq:kmeanspp}
\end{equation}
where $S$ is the set of already-selected centers.
We use a custom implementation built on top of scikit-learn~\citep{pedregosa2011scikit}, extended to support frozen centers and per-point weights.
Treatments selected in previous rounds are included as fixed (frozen) initial centers, so new selections are pushed away from already-observed regions.

The key difference between $k$-medoids and $k$-means++ is in their objective.
$k$-medoids minimizes the total distance between every data point and its closest medoid, which promotes both diversity and representativity of the selection.
$k$-means++ only considers distances between selected points, which emphasizes diversity over representativity: the selected set tends to be maximally spread out.
When the goal is to find top hits rather than uniformly cover the treatment space, diversity is preferable because it avoids selecting redundant treatments in dense but uninteresting regions.

\paragraph{Weighting strategies.}
We consider three weighting schemes for adaptive rounds:
(i) \emph{Delta weighting}: $w_i = \max(-\Delta_i, 0)$, where $\Delta_i = \hat{y}_i^{(\text{target})} - \hat{y}_i^{(\text{control})}$ is the predicted differential viability, so treatments predicted to be selectively effective receive higher weight.
(ii) \emph{Uncertainty weighting}: $w_i = \sigma_i$, where $\sigma_i$ is the MC dropout uncertainty~\citep{gal2016dropout} computed by enabling dropout (rate 0.1) on the Level~3 Sample Encoder and running 10 stochastic forward passes; the prediction is the mean and $\sigma_i$ is the standard deviation across passes.
(iii) \emph{Top-$k$}: instead of weighted clustering, we greedily select the $k$ treatments with the most negative $\Delta_i$ (top-$k$ delta) or highest $\sigma_i$ (top-$k$ uncertainty), without any diversity mechanism.

% ======================================================================
\FloatBarrier
\section{Baselines and hyperparameters}
\label{app:baselines}

\subsection{Comparison with related methods}

\cref{tab:baseline_comparison} provides a systematic comparison of drug response prediction methods from the literature.
Most existing methods require genomic or transcriptomic features to represent biological samples, which limits their applicability to patient-derived ex vivo models where such is often unavailable~\citep{pichotta2026pan}.
Monotherapy methods do not model drug interactions, and combination synergy methods typically predict a scalar synergy score rather than full dose-response surfaces.
Transfer learning approaches partially bridge the domain gap between cell lines and patient samples, but still require overlapping molecular features.
TabPFN~\citep{hollmann2025accurate}, XGBoost, and MLPs all meet the requirements for our setting (no genomics, dose-aware, combination-capable) when paired with a general drug representation such as Morgan fingerprints or learned embeddings; we include them as baselines.
TabPFN additionally shares the in-context learning paradigm with \method{}.
ComboLTR and PIICM produce full dose-response predictions for combinations without genomic features, but they lack a pretraining mechanism and learn drug representations from scratch on each sample.
This makes them effective when sufficient data is available per sample, but poorly suited to the few-shot regime where only a small number of observations are provided.

\begin{table}[h]
\caption{\textbf{Comparison of drug response prediction methods.} We list methods cited in related work and indicate their requirements and capabilities. \cmark{} indicates a desirable property. Methods that are not genomics- or transcriptomics-free require molecular profiling, which limits applicability to patient-derived ex vivo models. Methods without pretraining must be trained from scratch on each new dataset.}
\label{tab:baseline_comparison}
\centering
\small
\resizebox{\textwidth}{!}{
\begin{tabular}{lccccc}
\toprule
\textbf{Method} & \textbf{Combinations} & \textbf{Dose info} & \textbf{Pretrained} & \textbf{Genomics-free} & \textbf{Transcriptomics-free} \\
\midrule
\multicolumn{6}{l}{\emph{Monotherapy response prediction}} \\
\midrule
DNN \citep{sakellaropoulos2019deepDLDR}          & \xmark & \xmark & \xmark & \cmark & \xmark \\
tCNNs \citep{liu2019tcnns}                          & \xmark & \xmark & \xmark & \xmark & \cmark \\
DrugCell \citep{kuenzi2020predictingDLDR}            & \xmark & \xmark & \xmark & \xmark & \cmark \\
DeepCDR \citep{liu2020deepcdr}                       & \xmark & \xmark & \xmark & \xmark & \xmark \\
HiDRA \citep{jin2021hidra}                           & \xmark & \xmark & \xmark & \cmark & \xmark \\
GraphDRP \citep{nguyen2022graphdrp}                  & \xmark & \xmark & \xmark & \xmark & \cmark \\
XGDP \citep{wang2025drug}                            & \xmark & \xmark & \cmark & \cmark & \xmark \\
\midrule
\multicolumn{6}{l}{\emph{Combination synergy prediction}} \\
\midrule
DeepSynergy \citep{preuer2018deepsynergy}            & \cmark & \xmark & \xmark & \cmark & \xmark \\
ComboLTR \citep{julkunen2020comboltr}                & \cmark & \cmark & \xmark & \cmark & \cmark \\
PIICM \citep{ronneberg2024scalable}                  & \cmark & \cmark & \xmark & \cmark & \cmark \\
MatchMaker \citep{kuru2022matchmaker}                & \cmark & \xmark & \cmark & \cmark & \xmark \\
DeepDDS \citep{wang2022deepdds}                      & \cmark & \xmark & \xmark & \cmark & \xmark \\
MARSY \citep{elkhili2023marsy}                       & \cmark & \xmark & \xmark & \cmark & \xmark \\
DeepTraSynergy \citep{rafiei2023deeptrasynergy}      & \cmark & \xmark & \xmark & \cmark & \xmark \\
MGAE-DC \citep{zhang2023mgae}                        & \cmark & \xmark & \cmark & \cmark & \xmark \\
DconnC \citep{yan2024deep}                           & \cmark & \xmark & \xmark & \cmark & \xmark \\
PerturbSynX \citep{hafsath2025perturbsynx}           & \cmark & \cmark & \cmark & \cmark & \xmark \\
\midrule
\multicolumn{6}{l}{\emph{Transfer learning}} \\
\midrule
TRANSACT \citep{mourragui2021predictingTLDR}             & \xmark & \xmark & \cmark & \cmark & \xmark \\
TCRP \citep{ma2021fewshot}                   & \xmark & \xmark & \cmark & \xmark & \xmark \\
ScreenDL \citep{sederman2024screendlTLDR}            & \xmark & \xmark & \cmark & \cmark & \xmark \\
\midrule
\multicolumn{6}{l}{\emph{Foundation models}} \\
\midrule
PPC-FM \citep{pichotta2026pan}                         & \xmark & \cmark & \xmark & \cmark & \cmark \\
BAITSAO \citep{liu2025building}                       & \cmark & \xmark & \cmark & \cmark & \cmark \\
CancerGPT \citep{li2024cancergpt}                    & \cmark & \xmark & \xmark & \cmark & \cmark \\
SynerGPT \citep{edwardssynergpt}                     & \cmark & \xmark & \xmark & \cmark & \cmark \\
TabPFN \citep{hollmann2025accurate}                  & \cmark & \cmark & \cmark & \cmark & \cmark \\
\textbf{\method{} (ours)}                            & \cmark & \cmark & \cmark & \cmark & \cmark \\
\bottomrule
\end{tabular}
}
\end{table}

\subsection{Baseline hyperparameters}
\label{app:hyperparameters}

\paragraph{Input features for XGBoost and TabPFN.}
Both XGBoost and TabPFN use 512-dimensional Morgan fingerprints~\citep{rogers2010morganfingerprint} (radius~2) as drug representations.
For each drug in the combination, the fingerprint is concatenated with the corresponding dose value, yielding a $(512 + 1)$-dimensional vector per drug slot.
For combinations of $q$ drugs, the per-drug feature vectors are concatenated, producing a $513q$-dimensional input.
Empty drug slots (monotherapy observations in a combination dataset) are filled with zero vectors.

\paragraph{XGBoost.}
We use XGBoost with 100 estimators and squared error loss.
Hyperparameters are selected independently for each sample via 5-fold cross-validation with grid search over: max depth $\in \{2, 4, 6\}$, learning rate $\in \{0.05, 0.1, 0.3\}$, subsample ratio $\in \{0.7, 0.9, 1.0\}$, min child weight $\in \{1, 5, 10\}$, and L2 regularization $\lambda \in \{0.1, 1.0, 10.0\}$ (243 combinations).

\paragraph{TabPFN.}
We use TabPFN~\citep{hollmann2025accurate} with default settings and the same Morgan fingerprint input features as XGBoost.
TabPFN performs in-context learning: it conditions on the few-shot observations and predicts the held-out viability in a single forward pass, with no task-specific training.

\paragraph{MLP architecture.}
Both MLP baselines share the same architecture, selected through extensive manual tuning on pretraining validation loss.
The model uses learned embedding tables for drugs (vocabulary size matching the drug library) and samples, with embedding dimension $D{=}64$.
Doses are encoded via a linear projection from $\mathbb{R}^1$ to $\mathbb{R}^{64}$.
For each observation, the drug embedding and dose embedding are multiplied element-wise for each drug in the combination, then averaged across the $q$ drug slots to produce a single drug-dose representation in $\mathbb{R}^{64}$.
This is concatenated with the sample embedding to form a $128$-dimensional input, which is passed through 3 fully connected layers (hidden dimension~128, ReLU activations, dropout~0.2) followed by a linear output layer and sigmoid activation.

\paragraph{MLP (target-only).}
Trained from scratch on each target sample using Adam (lr $= 10^{-3}$, no weight decay), gradient clipping at 1.0, batch size 32, for up to 200 epochs.
Each epoch consists of 100 iterations (sampling with replacement from the few-shot context).
We use a 20\% validation split for early stopping and select the checkpoint with the lowest validation loss.

\paragraph{MLP (fine-tuned).}
Same architecture, pretrained on the same 40-dataset corpus as \method{} (excluding the held-out evaluation dataset) using Adam (lr $= 10^{-3}$, no weight decay), batch size 128, gradient clipping at 1.0, for 50 epochs over the full corpus (${\sim}$234K iterations per epoch, ${\sim}$11.7M total training steps).
A 5\% held-out split of (sample, drug) pairs is used for validation, and the checkpoint with the lowest validation loss is selected.
The pretrained model uses the same drug library as \method{} for drug identity encoding.
At evaluation time, a new sample embedding is initialized for the target sample, and the full model is fine-tuned on the few-shot context with a learning rate of $10^{-4}$ for up to 200 epochs with early stopping on a 20\% validation split.

% ======================================================================
\FloatBarrier
\section{Few-shot additional results}
\label{app:few_shot_additional}
\label{app:computation_time}

\begin{figure}[htbp]
    \centering
    \includegraphics[width=\textwidth]{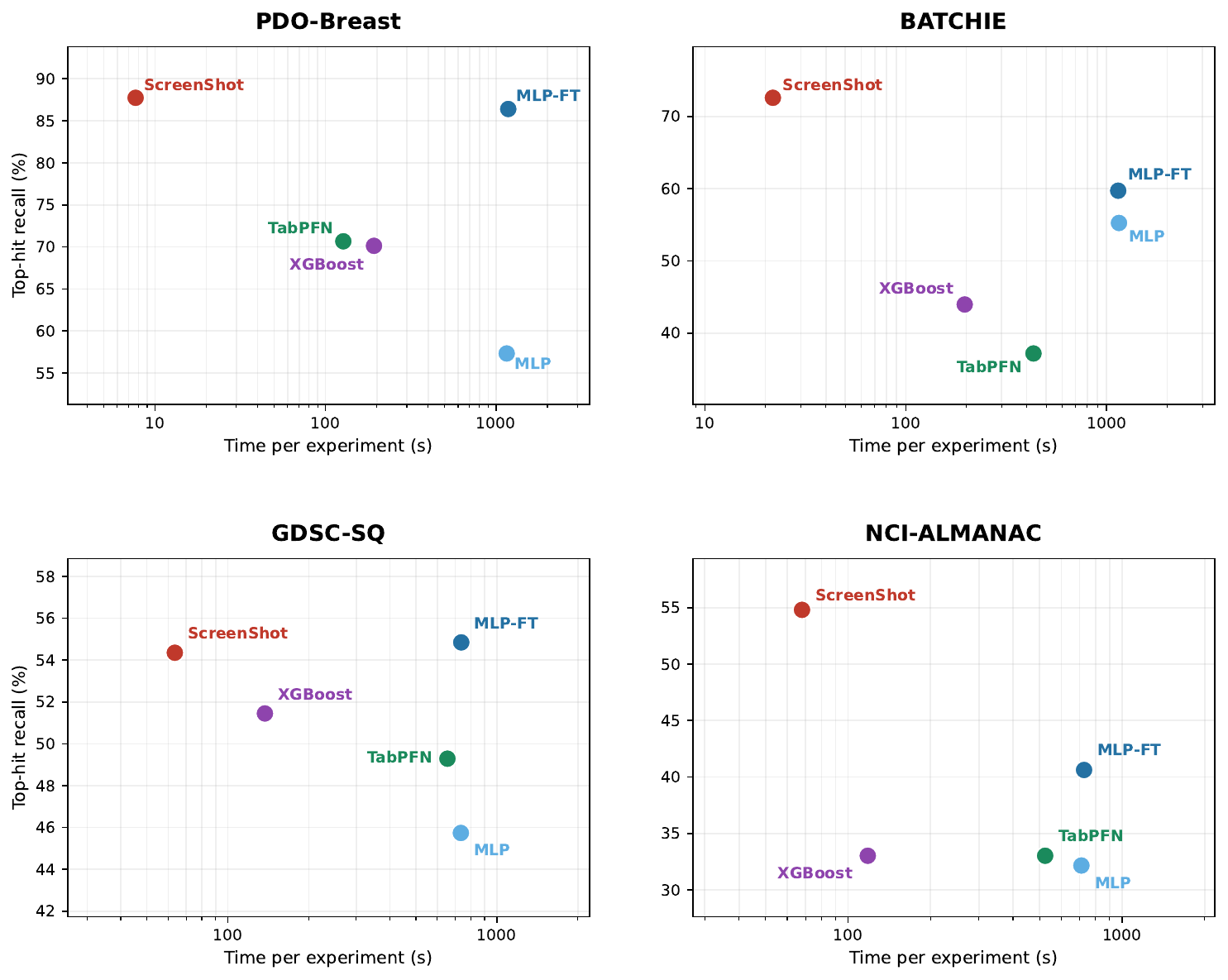}
    \caption{Top-hit recall vs.\ computation time per experiment at 100-shot budget. ScreenShot (red) is consistently the fastest and most accurate method across all four datasets. All GPU methods ran on NVIDIA A100 80GB GPUs; XGBoost ran on 8 CPU cores.}
    \label{fig:time_vs_accuracy}
\end{figure}

\cref{fig:time_vs_accuracy} shows the trade-off between top-hit recall and computation time for each method at a budget of 100 shots.
ScreenShot achieves the highest accuracy on every dataset while being 5--50$\times$ faster than all baselines.
MLP and MLP-FT require training on the few-shot context for each experiment, resulting in computation times of 10--20 minutes per experiment.
XGBoost runs on CPU (8 cores) and takes 2--3 minutes.
TabPFN is slower than ScreenShot despite also being inference-only, due to its larger model size and its use of self-attention across both context observations and input features.
ScreenShot requires only a single forward pass through the pretrained model, completing in 8--68 seconds depending on dataset size.

% ======================================================================
\FloatBarrier
\section{Active learning additional results}
\label{app:al_additional}

\subsection{Hit recall across datasets}
\label{app:al_recall}

\cref{tab:al_hit_recall} reports top-hit recall for the four experimental design strategies across three datasets: random selection, $k$-medoids cold start (100\% budget), and two adaptive strategies with 33\% cold start followed by 1 or 2 delta-weighted $k$-means++ rounds.
Adaptive methods consistently outperform random selection and cold-start baselines.
On BATCHIE, adaptive selection with two rounds reaches 84.8\% at budget 300, compared to 75.8\% for random, a gain of 9 percentage points.
On GDSC-SQ and NCI-ALMANAC, adaptive methods improve by 4--7 points over random.

\begin{table}[h]
\caption{\textbf{Active learning: hit recall (\%).} Mean$\pm$std across 5 seeds. \textbf{Bold}: best per column.}
\label{tab:al_hit_recall}
\centering
\small
\resizebox{\textwidth}{!}{
\begin{tabular}{lrrrrrrrrr}
\toprule
\textbf{Method} & \multicolumn{3}{c}{\textbf{BATCHIE}} & \multicolumn{3}{c}{\textbf{GDSC-SQ}} & \multicolumn{3}{c}{\textbf{NCI-ALMANAC}} \\
\cmidrule(lr){2-4}
\cmidrule(lr){5-7}
\cmidrule(lr){8-10}
 & 100 & 200 & 300 & 100 & 200 & 300 & 100 & 200 & 300 \\
\midrule
Random & 71.6{\scriptsize$\pm$1.6} & 74.4{\scriptsize$\pm$1.7} & 75.8{\scriptsize$\pm$1.3} & 56.3{\scriptsize$\pm$2.0} & 60.6{\scriptsize$\pm$1.8} & 63.0{\scriptsize$\pm$1.2} & 49.4{\scriptsize$\pm$2.0} & 54.0{\scriptsize$\pm$0.6} & 56.6{\scriptsize$\pm$0.6} \\
Cold-start ($k$-medoids) & 72.9{\scriptsize$\pm$1.4} & 75.1{\scriptsize$\pm$1.1} & 76.1{\scriptsize$\pm$0.7} & 60.4{\scriptsize$\pm$1.7} & 63.0{\scriptsize$\pm$1.4} & 65.2{\scriptsize$\pm$1.1} & \textbf{60.3{\scriptsize$\pm$0.8}} & 60.8{\scriptsize$\pm$1.4} & 61.2{\scriptsize$\pm$1.5} \\
Adaptive (1 round) & 77.3{\scriptsize$\pm$1.4} & 81.2{\scriptsize$\pm$0.8} & 84.0{\scriptsize$\pm$0.7} & \textbf{61.1{\scriptsize$\pm$0.6}} & \textbf{65.5{\scriptsize$\pm$0.8}} & 67.0{\scriptsize$\pm$1.6} & 57.6{\scriptsize$\pm$2.9} & \textbf{62.3{\scriptsize$\pm$0.9}} & 62.9{\scriptsize$\pm$1.0} \\
Adaptive (2 rounds) & \textbf{77.5{\scriptsize$\pm$0.9}} & \textbf{83.0{\scriptsize$\pm$2.6}} & \textbf{84.8{\scriptsize$\pm$1.9}} & 59.9{\scriptsize$\pm$1.4} & 65.3{\scriptsize$\pm$0.7} & \textbf{67.4{\scriptsize$\pm$1.0}} & 58.2{\scriptsize$\pm$2.0} & 61.5{\scriptsize$\pm$0.9} & \textbf{63.6{\scriptsize$\pm$1.1}} \\
\bottomrule
\end{tabular}
}
\end{table}

\subsection{Effect of cold-start proportion and number of rounds}
\label{app:rounds}

\cref{tab:ablation_rounds} examines how the split between cold-start and adaptive budget affects performance on the BATCHIE combination screen.
The 33\% cold-start split achieves the best hit recall (84.8\% at budget~300), while allocating more to cold start (50\%) gives the best MAE and Pearson correlation.
This suggests a balance between initial coverage and adaptive exploitation.
Two rounds consistently outperform one round, though the marginal gain is modest (${\sim}$0.5--1 percentage points).
All adaptive strategies substantially outperform pure $k$-medoids cold start (76.1\%) in hit recall.
However, pure $k$-medoids achieves the best MAE, which is consistent with the known effectiveness of $k$-medoids for reducing global prediction error through its diversity-representativity tradeoff~\citep{dediscrepancy}, but demonstrates that minimizing global error does not directly translate to better top-hit detection.

\begin{table}[htbp]
\caption{\textbf{Effect of cold-start proportion and number of adaptive rounds} on the BATCHIE combination screen. All adaptive methods use the proposed strategy ($k$-medoids cold start, $k$-means++ with delta weighting on final embeddings). Bold indicates best per column.}
\label{tab:ablation_rounds}
\centering
\small
\resizebox{\columnwidth}{!}{%
\begin{tabular}{lrrrrrr}
\toprule
\textbf{Method} & \multicolumn{3}{c}{\textbf{Hit Recall (\%)}} & \multicolumn{3}{c}{\textbf{MAE}} \\
\cmidrule(lr){2-4}
\cmidrule(lr){5-7}
 & 100 & 200 & 300 & 100 & 200 & 300 \\
\midrule
$k$-medoids (100\% cold start) & 72.9{\scriptsize$\pm$1.4} & 75.1{\scriptsize$\pm$1.1} & 76.1{\scriptsize$\pm$0.7} & \textbf{0.107{\scriptsize$\pm$0.001}} & \textbf{0.097{\scriptsize$\pm$0.001}} & \textbf{0.093{\scriptsize$\pm$0.001}} \\
Adaptive (50\% CS, 1 round) & 75.3{\scriptsize$\pm$1.4} & 78.9{\scriptsize$\pm$1.6} & 82.2{\scriptsize$\pm$1.3} & 0.109{\scriptsize$\pm$0.001} & 0.099{\scriptsize$\pm$0.001} & \textbf{0.093{\scriptsize$\pm$0.001}} \\
Adaptive (50\% CS, 2 rounds) & 75.8{\scriptsize$\pm$0.8} & 80.5{\scriptsize$\pm$1.4} & 82.6{\scriptsize$\pm$1.2} & 0.109{\scriptsize$\pm$0.001} & 0.098{\scriptsize$\pm$0.001} & \textbf{0.093{\scriptsize$\pm$0.001}} \\
Adaptive (33\% CS, 1 round) & 77.3{\scriptsize$\pm$1.4} & 81.2{\scriptsize$\pm$0.8} & 84.0{\scriptsize$\pm$0.7} & 0.110{\scriptsize$\pm$0.002} & 0.100{\scriptsize$\pm$0.001} & 0.095{\scriptsize$\pm$0.001} \\
Adaptive (33\% CS, 2 rounds) & \textbf{77.5{\scriptsize$\pm$0.9}} & \textbf{83.0{\scriptsize$\pm$2.6}} & \textbf{84.8{\scriptsize$\pm$1.9}} & 0.110{\scriptsize$\pm$0.001} & 0.100{\scriptsize$\pm$0.001} & 0.094{\scriptsize$\pm$0.001} \\
Adaptive (20\% CS, 1 round) & 74.7{\scriptsize$\pm$0.9} & 79.1{\scriptsize$\pm$1.0} & 83.3{\scriptsize$\pm$1.7} & 0.112{\scriptsize$\pm$0.001} & 0.104{\scriptsize$\pm$0.001} & 0.099{\scriptsize$\pm$0.001} \\
Adaptive (20\% CS, 2 rounds) & 75.5{\scriptsize$\pm$1.9} & 80.0{\scriptsize$\pm$1.4} & 83.8{\scriptsize$\pm$2.0} & 0.112{\scriptsize$\pm$0.001} & 0.104{\scriptsize$\pm$0.001} & 0.098{\scriptsize$\pm$0.001} \\
\bottomrule
\end{tabular}}
\end{table}

% ======================================================================
\FloatBarrier
\section{Architecture ablation}
\label{app:arch_ablation}

\cref{tab:arch_ablation} compares the full ScreenShot architecture against three ablated variants on the BATCHIE and PDO-Breast datasets.
\emph{No Level~3} removes the sample-level self-attention that aggregates across perturbations.
\emph{Flat attention} replaces the hierarchical encoder with a single self-attention layer over all flattened context observations followed by cross-attention.
\emph{No Level~1~\&~3} removes both the drug-combination encoder and the sample-level aggregation; drug and dose embeddings are combined via element-wise product and averaged across drugs in the combination (as in the MLP baseline), producing a single vector per observation that is passed directly to the cross-attention response encoder.

\begin{figure}[htbp]
    \centering
    \includegraphics[width=\textwidth]{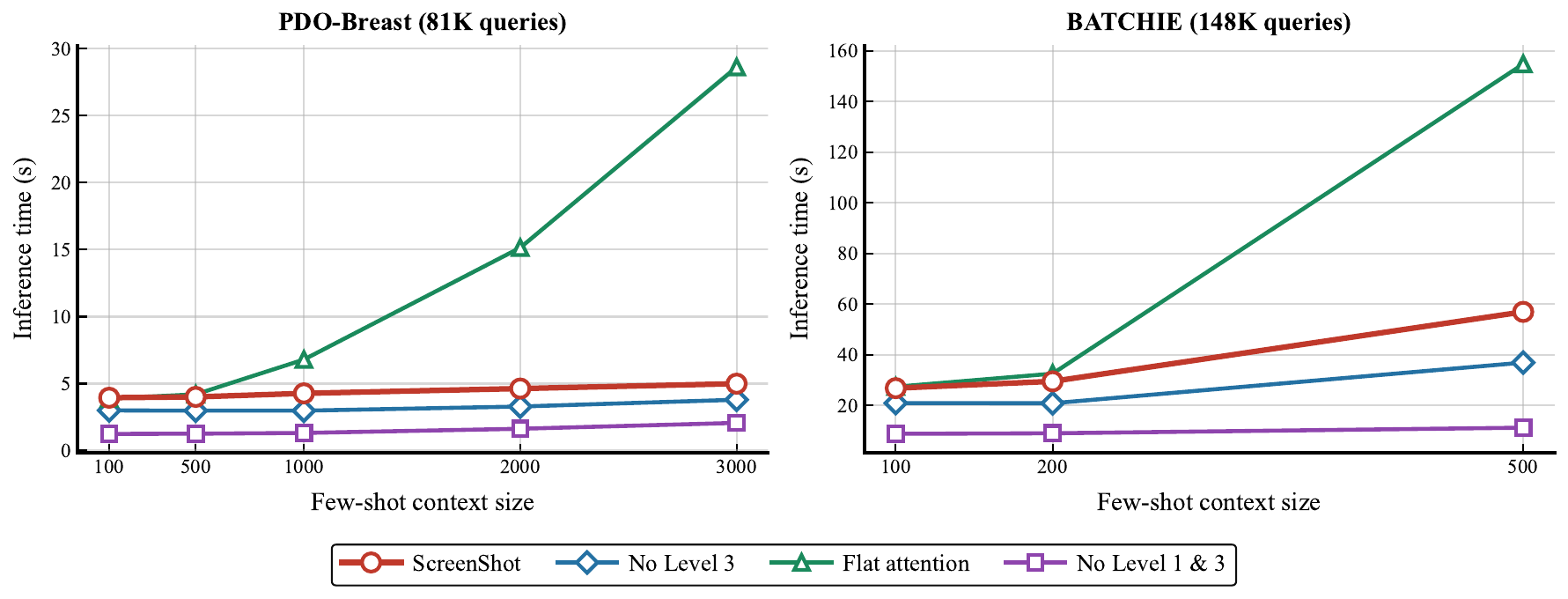}
    \caption{Inference time vs.\ context size on PDO-Breast (left) and BATCHIE (right), measured on NVIDIA H100 GPU.}
    \label{fig:inference_time_ablation}
\end{figure}

The full architecture achieves the best hit recall on 9 out of 10 dataset/budget combinations.
\emph{No Level~1~\&~3}, which reduces the model to the cross-attention level, drops significantly on BATCHIE (54.7\% vs.\ 76.0\% at budget~300) and saturates on PDO-Breast (${\sim}$78\% regardless of budget), indicating that a single encoder level lacks the capacity to capture the structure of screening data.
\emph{Flat attention} and \emph{No Level~3} perform closer to the full model but consistently trail on BATCHIE, where the combinatorial treatment space is larger.
Overall, each level contributes to prediction quality at the cost of a reasonable increase in computation (\cref{fig:inference_time_ablation}).

\begin{table*}[htbp]
\centering
\small
\caption{Architecture ablation: top-hit recall (\%) across few-shot budgets on BATCHIE and PDO-Breast.
Values: mean$\pm$std across 5 seeds. \textbf{Bold}: best per column.}
\label{tab:arch_ablation}
\vspace{-4pt}
\providecommand{\tpm}[1]{{\tiny$\pm$#1}}
\resizebox{\textwidth}{!}{%
\setlength{\tabcolsep}{4pt}
\begin{tabular}{lcccccccccc}
\toprule
 & \multicolumn{5}{c}{\textsc{BATCHIE}} & \multicolumn{5}{c}{\textsc{PDO-Breast}} \\
\cmidrule(lr){2-6} \cmidrule(lr){7-11}
 & 10 & 50 & 100 & 200 & 300 & 10 & 50 & 100 & 200 & 300 \\
\midrule
ScreenShot & \bf 63.9\tpm{3.1} & \bf 69.0\tpm{2.4} & \bf 72.6\tpm{3.0} & \bf 75.0\tpm{2.5} & \bf 76.0\tpm{1.4} & \bf 78.7\tpm{2.5} & 81.9\tpm{2.2} & \bf 87.7\tpm{2.4} & \bf 93.6\tpm{1.7} & \bf 96.0\tpm{1.6} \\
Flat attention & 60.3\tpm{2.2} & 65.6\tpm{1.6} & 68.6\tpm{0.6} & 70.9\tpm{0.9} & 72.3\tpm{1.3} & 76.3\tpm{6.5} & 82.9\tpm{3.0} & 86.9\tpm{3.2} & 91.5\tpm{2.6} & 94.7\tpm{1.6} \\
No Level 3 & 59.5\tpm{2.0} & 66.9\tpm{1.2} & 70.9\tpm{1.3} & 74.5\tpm{1.1} & 75.6\tpm{1.0} & 75.5\tpm{3.8} & \bf 83.5\tpm{1.8} & 87.7\tpm{3.2} & 92.0\tpm{2.1} & 95.2\tpm{3.1} \\
No Level 1 \& 3 & 51.0\tpm{3.4} & 53.2\tpm{1.5} & 55.0\tpm{1.5} & 54.4\tpm{0.4} & 54.7\tpm{0.9} & 77.1\tpm{1.7} & 78.4\tpm{2.0} & 77.9\tpm{1.2} & 78.1\tpm{1.5} & 78.4\tpm{1.1} \\
\bottomrule
\end{tabular}}
\end{table*}

\cref{fig:inference_time_ablation} shows inference time as a function of context size on both PDO-Breast and BATCHIE.
On PDO-Breast, ScreenShot scales from 3.9\,s at 100 context observations to 5.0\,s at 3{,}000, because the dominant self-attention at Level~3 operates over perturbations ($p \ll n$) rather than individual observations.
The flat attention variant exhibits quadratic scaling, reaching 28.6\,s at 3{,}000 observations (5.7$\times$ slower).
On BATCHIE, flat attention reaches 155\,s at 500 context observations (2.7$\times$ slower than ScreenShot at the same budget).
This confirms the computational motivation for the hierarchical design discussed in \cref{sec:architecture}.

% ======================================================================
\FloatBarrier
\section{Active learning ablation}
\label{app:ablation}

We ablate each component of the adaptive experimental design strategy described in \cref{sec:experiments}.
The \emph{proposed} method uses $k$-medoids on drug-dose embeddings for cold start (33\% of budget), followed by $k$-means++ on final model embeddings with delta weighting adaptive rounds.
Each ablation modifies exactly one component.
We evaluate on three datasets (BATCHIE, GDSC2, NCI-ALMANAC) with budgets of 100, 200, and 300 treatments per sample, averaged over 5 seeds. The ``Uncertainty weighting'' ablation uses MC dropout uncertainty instead of predicted delta as selection weights.
Uncertainty is computed by enabling dropout (applied to the Level~3 Sample Encoder only) and running 10 stochastic forward passes.
The prediction is the mean across passes; uncertainty is the standard deviation.

\cref{tab:ablation_hit_recall} reports hit recall for each variant.
\cref{tab:ablation_mae_pearson} reports MAE and Pearson correlation, which measure overall prediction quality rather than hit detection.

\begin{table}[h]
\caption{\textbf{Active learning ablation: Hit Recall@20\%.} The first row reports absolute hit recall (\%); all other rows report the difference in percentage points relative to the proposed method. Each row removes or replaces one component. All methods use 33\% cold start budget. \colorbox{better1}{Blue}: significantly better than proposed; \colorbox{worse1}{red}: significantly worse (Wilcoxon signed-rank test). Shading intensity: light $p < 0.05$, medium $p < 0.01$, dark $p < 0.001$.}
\label{tab:ablation_hit_recall}
\centering
\small
\resizebox{\textwidth}{!}{
\begin{tabular}{lrrrrrrrrr}
\toprule
\textbf{Method} & \multicolumn{3}{c}{\textbf{BATCHIE}} & \multicolumn{3}{c}{\textbf{GDSC2}} & \multicolumn{3}{c}{\textbf{NCI-ALMANAC}} \\
\cmidrule(lr){2-4}
\cmidrule(lr){5-7}
\cmidrule(lr){8-10}
 & 100 & 200 & 300 & 100 & 200 & 300 & 100 & 200 & 300 \\
\midrule
\multicolumn{10}{l}{\emph{1 adaptive round}} \\
\midrule
Proposed & 77.3{\scriptsize$\pm$1.4} & 81.2{\scriptsize$\pm$0.8} & 84.0{\scriptsize$\pm$0.7} & 61.1{\scriptsize$\pm$0.6} & 65.5{\scriptsize$\pm$0.8} & 67.0{\scriptsize$\pm$1.6} & 57.6{\scriptsize$\pm$2.9} & 62.3{\scriptsize$\pm$0.9} & 62.9{\scriptsize$\pm$1.0} \\
No delta weighting & \cellcolor{worse3}-4.3{\scriptsize$\pm$1.0} & \cellcolor{worse3}-4.9{\scriptsize$\pm$0.9} & \cellcolor{worse3}-5.8{\scriptsize$\pm$1.3} & \cellcolor{worse3}-4.0{\scriptsize$\pm$1.0} & \cellcolor{worse3}-4.3{\scriptsize$\pm$0.6} & \cellcolor{worse3}-3.0{\scriptsize$\pm$0.6} & -0.7{\scriptsize$\pm$0.8} & \cellcolor{worse1}-2.0{\scriptsize$\pm$0.8} & \cellcolor{worse3}-1.8{\scriptsize$\pm$0.5} \\
No $k$-means++ & \cellcolor{worse3}-3.2{\scriptsize$\pm$0.6} & \cellcolor{worse3}-3.1{\scriptsize$\pm$0.6} & \cellcolor{worse3}-4.3{\scriptsize$\pm$0.8} & +0.5{\scriptsize$\pm$0.6} & -0.6{\scriptsize$\pm$0.5} & -0.8{\scriptsize$\pm$0.5} & +0.8{\scriptsize$\pm$0.7} & -0.3{\scriptsize$\pm$0.5} & +0.1{\scriptsize$\pm$0.5} \\
No final embeddings & \cellcolor{worse3}-2.8{\scriptsize$\pm$0.7} & -0.4{\scriptsize$\pm$0.5} & \cellcolor{worse2}-1.3{\scriptsize$\pm$0.6} & +0.1{\scriptsize$\pm$0.9} & -0.8{\scriptsize$\pm$0.7} & +0.2{\scriptsize$\pm$0.6} & +1.3{\scriptsize$\pm$0.7} & -0.9{\scriptsize$\pm$0.5} & -0.1{\scriptsize$\pm$0.4} \\
$k$-means++ cold start & -1.7{\scriptsize$\pm$0.9} & +0.1{\scriptsize$\pm$1.0} & +0.2{\scriptsize$\pm$0.7} & -0.3{\scriptsize$\pm$0.9} & -0.6{\scriptsize$\pm$0.8} & +0.2{\scriptsize$\pm$0.7} & \cellcolor{worse2}-3.1{\scriptsize$\pm$1.0} & \cellcolor{worse1}-1.6{\scriptsize$\pm$0.7} & +0.0{\scriptsize$\pm$0.6} \\
Random cold start & -0.4{\scriptsize$\pm$0.9} & -0.0{\scriptsize$\pm$0.9} & +0.0{\scriptsize$\pm$1.0} & \cellcolor{worse1}-1.4{\scriptsize$\pm$0.8} & -1.2{\scriptsize$\pm$0.9} & +0.0{\scriptsize$\pm$0.7} & \cellcolor{worse3}-4.1{\scriptsize$\pm$1.0} & \cellcolor{worse3}-4.7{\scriptsize$\pm$1.1} & \cellcolor{worse2}-1.7{\scriptsize$\pm$0.7} \\
Uncertainty weighting & \cellcolor{worse3}-4.3{\scriptsize$\pm$0.9} & \cellcolor{worse3}-4.3{\scriptsize$\pm$0.9} & \cellcolor{worse3}-5.4{\scriptsize$\pm$0.9} & \cellcolor{worse1}-1.8{\scriptsize$\pm$0.6} & \cellcolor{worse2}-2.5{\scriptsize$\pm$0.7} & \cellcolor{worse1}-2.0{\scriptsize$\pm$0.6} & +0.8{\scriptsize$\pm$1.0} & -0.8{\scriptsize$\pm$0.7} & -0.5{\scriptsize$\pm$0.5} \\
Top-$k$ delta & \cellcolor{worse3}-5.8{\scriptsize$\pm$0.8} & \cellcolor{worse3}-3.4{\scriptsize$\pm$0.5} & \cellcolor{worse3}-3.4{\scriptsize$\pm$0.6} & \cellcolor{worse3}-10.0{\scriptsize$\pm$0.9} & \cellcolor{worse3}-8.1{\scriptsize$\pm$0.8} & \cellcolor{worse3}-7.0{\scriptsize$\pm$0.5} & \cellcolor{worse3}-7.4{\scriptsize$\pm$1.1} & \cellcolor{worse3}-7.7{\scriptsize$\pm$1.4} & \cellcolor{worse3}-11.8{\scriptsize$\pm$1.4} \\
Top-$k$ uncertainty & \cellcolor{worse3}-8.1{\scriptsize$\pm$1.3} & \cellcolor{worse3}-5.6{\scriptsize$\pm$0.9} & \cellcolor{worse3}-6.0{\scriptsize$\pm$0.9} & \cellcolor{worse3}-6.6{\scriptsize$\pm$1.0} & \cellcolor{worse3}-6.4{\scriptsize$\pm$0.8} & \cellcolor{worse3}-5.7{\scriptsize$\pm$0.8} & \cellcolor{worse3}-6.4{\scriptsize$\pm$1.0} & \cellcolor{worse3}-7.4{\scriptsize$\pm$1.6} & \cellcolor{worse3}-9.6{\scriptsize$\pm$1.7} \\
\midrule
\multicolumn{10}{l}{\emph{2 adaptive rounds}} \\
\midrule
Proposed & 77.5{\scriptsize$\pm$0.9} & 83.0{\scriptsize$\pm$2.6} & 84.8{\scriptsize$\pm$1.9} & 59.9{\scriptsize$\pm$1.4} & 65.3{\scriptsize$\pm$0.7} & 67.4{\scriptsize$\pm$1.0} & 58.2{\scriptsize$\pm$2.0} & 61.5{\scriptsize$\pm$0.9} & 63.6{\scriptsize$\pm$1.1} \\
No delta weighting & \cellcolor{worse3}-5.2{\scriptsize$\pm$1.0} & \cellcolor{worse3}-7.0{\scriptsize$\pm$1.2} & \cellcolor{worse3}-5.7{\scriptsize$\pm$1.2} & \cellcolor{worse3}-2.9{\scriptsize$\pm$0.8} & \cellcolor{worse3}-4.4{\scriptsize$\pm$0.6} & \cellcolor{worse3}-3.2{\scriptsize$\pm$0.7} & \cellcolor{worse3}-3.0{\scriptsize$\pm$0.8} & -1.1{\scriptsize$\pm$0.7} & \cellcolor{worse3}-2.4{\scriptsize$\pm$0.5} \\
No $k$-means++ & \cellcolor{worse2}-2.6{\scriptsize$\pm$0.9} & \cellcolor{worse3}-3.9{\scriptsize$\pm$0.8} & \cellcolor{worse3}-4.0{\scriptsize$\pm$0.7} & +0.9{\scriptsize$\pm$0.7} & -0.7{\scriptsize$\pm$0.8} & -0.2{\scriptsize$\pm$0.6} & -0.0{\scriptsize$\pm$0.7} & -0.2{\scriptsize$\pm$0.5} & -0.2{\scriptsize$\pm$0.4} \\
No final embeddings & \cellcolor{worse3}-3.9{\scriptsize$\pm$1.0} & \cellcolor{worse2}-3.1{\scriptsize$\pm$0.9} & \cellcolor{worse1}-1.8{\scriptsize$\pm$0.7} & \cellcolor{better1}+2.3{\scriptsize$\pm$0.9} & +0.9{\scriptsize$\pm$0.7} & +0.7{\scriptsize$\pm$0.6} & +0.4{\scriptsize$\pm$0.6} & +0.1{\scriptsize$\pm$0.6} & -0.1{\scriptsize$\pm$0.5} \\
$k$-means++ cold start & -0.1{\scriptsize$\pm$0.9} & -0.9{\scriptsize$\pm$1.1} & -0.2{\scriptsize$\pm$0.9} & -0.1{\scriptsize$\pm$1.0} & -0.1{\scriptsize$\pm$0.8} & -0.4{\scriptsize$\pm$0.6} & \cellcolor{worse3}-3.8{\scriptsize$\pm$1.0} & -0.7{\scriptsize$\pm$0.8} & -0.8{\scriptsize$\pm$0.6} \\
Random cold start & -0.4{\scriptsize$\pm$0.7} & -1.2{\scriptsize$\pm$0.9} & +0.4{\scriptsize$\pm$1.0} & -1.5{\scriptsize$\pm$1.2} & -0.8{\scriptsize$\pm$0.8} & -0.7{\scriptsize$\pm$0.4} & \cellcolor{worse3}-3.8{\scriptsize$\pm$1.0} & \cellcolor{worse1}-1.7{\scriptsize$\pm$0.8} & \cellcolor{worse1}-1.5{\scriptsize$\pm$0.6} \\
Uncertainty weighting & \cellcolor{worse2}-3.1{\scriptsize$\pm$0.9} & \cellcolor{worse3}-5.6{\scriptsize$\pm$1.0} & \cellcolor{worse3}-6.2{\scriptsize$\pm$0.9} & \cellcolor{worse1}-1.9{\scriptsize$\pm$0.8} & \cellcolor{worse3}-3.8{\scriptsize$\pm$0.7} & \cellcolor{worse2}-2.4{\scriptsize$\pm$0.8} & -0.5{\scriptsize$\pm$0.8} & -0.7{\scriptsize$\pm$0.5} & -0.9{\scriptsize$\pm$0.5} \\
Top-$k$ delta & \cellcolor{worse3}-6.1{\scriptsize$\pm$1.2} & \cellcolor{worse3}-5.6{\scriptsize$\pm$0.8} & \cellcolor{worse3}-4.6{\scriptsize$\pm$0.8} & \cellcolor{worse3}-6.1{\scriptsize$\pm$1.1} & \cellcolor{worse3}-6.7{\scriptsize$\pm$0.8} & \cellcolor{worse3}-6.0{\scriptsize$\pm$0.8} & \cellcolor{worse3}-6.6{\scriptsize$\pm$0.8} & \cellcolor{worse3}-7.3{\scriptsize$\pm$1.6} & \cellcolor{worse3}-10.9{\scriptsize$\pm$1.9} \\
Top-$k$ uncertainty & \cellcolor{worse3}-8.5{\scriptsize$\pm$1.4} & \cellcolor{worse3}-7.8{\scriptsize$\pm$1.0} & \cellcolor{worse3}-8.0{\scriptsize$\pm$1.1} & \cellcolor{worse3}-4.3{\scriptsize$\pm$0.9} & \cellcolor{worse3}-5.4{\scriptsize$\pm$0.8} & \cellcolor{worse3}-5.7{\scriptsize$\pm$0.8} & \cellcolor{worse3}-5.0{\scriptsize$\pm$0.7} & \cellcolor{worse3}-4.2{\scriptsize$\pm$0.8} & \cellcolor{worse3}-4.3{\scriptsize$\pm$0.8} \\
\bottomrule
\end{tabular}
}
\end{table}

\begin{table}[h]
\caption{\textbf{Active learning ablation: MAE and Pearson correlation.} Same ablation as Table~\ref{tab:ablation_hit_recall}. Bold indicates best per column (lowest MAE, highest Pearson).}
\label{tab:ablation_mae_pearson}
\centering
\small
\resizebox{\textwidth}{!}{
\begin{tabular}{lrrrrrrrrr}
\toprule
\textbf{Method} & \multicolumn{3}{c}{\textbf{BATCHIE}} & \multicolumn{3}{c}{\textbf{GDSC2}} & \multicolumn{3}{c}{\textbf{NCI-ALMANAC}} \\
\cmidrule(lr){2-4}
\cmidrule(lr){5-7}
\cmidrule(lr){8-10}
 & 100 & 200 & 300 & 100 & 200 & 300 & 100 & 200 & 300 \\
\midrule
\multicolumn{10}{l}{\emph{MAE --- 1 adaptive round}} \\
\midrule
Proposed & 0.110{\scriptsize$\pm$0.002} & 0.100{\scriptsize$\pm$0.001} & 0.095{\scriptsize$\pm$0.001} & 0.132{\scriptsize$\pm$0.001} & 0.126{\scriptsize$\pm$0.001} & 0.125{\scriptsize$\pm$0.001} & 0.144{\scriptsize$\pm$0.003} & 0.134{\scriptsize$\pm$0.001} & 0.131{\scriptsize$\pm$0.001} \\
No delta weighting & 0.111{\scriptsize$\pm$0.001} & 0.100{\scriptsize$\pm$0.001} & 0.095{\scriptsize$\pm$0.001} & 0.132{\scriptsize$\pm$0.001} & 0.126{\scriptsize$\pm$0.001} & 0.123{\scriptsize$\pm$0.001} & 0.147{\scriptsize$\pm$0.003} & 0.137{\scriptsize$\pm$0.001} & 0.132{\scriptsize$\pm$0.001} \\
No $k$-means++ & 0.112{\scriptsize$\pm$0.001} & 0.102{\scriptsize$\pm$0.001} & 0.096{\scriptsize$\pm$0.000} & \textbf{0.131{\scriptsize$\pm$0.001}} & 0.127{\scriptsize$\pm$0.000} & 0.126{\scriptsize$\pm$0.001} & 0.144{\scriptsize$\pm$0.002} & 0.136{\scriptsize$\pm$0.002} & 0.132{\scriptsize$\pm$0.001} \\
No final embeddings & 0.113{\scriptsize$\pm$0.001} & 0.102{\scriptsize$\pm$0.001} & 0.097{\scriptsize$\pm$0.000} & 0.133{\scriptsize$\pm$0.001} & 0.130{\scriptsize$\pm$0.001} & 0.128{\scriptsize$\pm$0.001} & \textbf{0.142{\scriptsize$\pm$0.002}} & 0.136{\scriptsize$\pm$0.001} & 0.132{\scriptsize$\pm$0.001} \\
$k$-means++ cold start & 0.113{\scriptsize$\pm$0.001} & 0.102{\scriptsize$\pm$0.001} & 0.097{\scriptsize$\pm$0.000} & 0.133{\scriptsize$\pm$0.002} & 0.128{\scriptsize$\pm$0.001} & 0.125{\scriptsize$\pm$0.001} & 0.145{\scriptsize$\pm$0.002} & 0.137{\scriptsize$\pm$0.001} & 0.132{\scriptsize$\pm$0.001} \\
Random cold start & 0.113{\scriptsize$\pm$0.002} & 0.105{\scriptsize$\pm$0.001} & 0.100{\scriptsize$\pm$0.001} & 0.135{\scriptsize$\pm$0.001} & 0.130{\scriptsize$\pm$0.001} & 0.126{\scriptsize$\pm$0.001} & 0.146{\scriptsize$\pm$0.001} & 0.138{\scriptsize$\pm$0.001} & 0.133{\scriptsize$\pm$0.001} \\
Uncertainty weighting & \textbf{0.108{\scriptsize$\pm$0.000}} & \textbf{0.097{\scriptsize$\pm$0.000}} & \textbf{0.092{\scriptsize$\pm$0.001}} & 0.131{\scriptsize$\pm$0.001} & \textbf{0.125{\scriptsize$\pm$0.001}} & \textbf{0.122{\scriptsize$\pm$0.001}} & 0.142{\scriptsize$\pm$0.001} & \textbf{0.132{\scriptsize$\pm$0.001}} & \textbf{0.128{\scriptsize$\pm$0.001}} \\
Top-$k$ delta & 0.124{\scriptsize$\pm$0.001} & 0.111{\scriptsize$\pm$0.001} & 0.104{\scriptsize$\pm$0.001} & 0.144{\scriptsize$\pm$0.003} & 0.138{\scriptsize$\pm$0.001} & 0.137{\scriptsize$\pm$0.000} & 0.165{\scriptsize$\pm$0.005} & 0.152{\scriptsize$\pm$0.004} & 0.156{\scriptsize$\pm$0.003} \\
Top-$k$ uncertainty & 0.121{\scriptsize$\pm$0.001} & 0.108{\scriptsize$\pm$0.001} & 0.102{\scriptsize$\pm$0.000} & 0.141{\scriptsize$\pm$0.003} & 0.133{\scriptsize$\pm$0.002} & 0.129{\scriptsize$\pm$0.001} & 0.157{\scriptsize$\pm$0.002} & 0.148{\scriptsize$\pm$0.003} & 0.146{\scriptsize$\pm$0.002} \\
\midrule
\multicolumn{10}{l}{\emph{MAE --- 2 adaptive rounds}} \\
\midrule
Proposed & 0.110{\scriptsize$\pm$0.001} & 0.100{\scriptsize$\pm$0.001} & 0.094{\scriptsize$\pm$0.001} & 0.132{\scriptsize$\pm$0.002} & 0.126{\scriptsize$\pm$0.001} & 0.124{\scriptsize$\pm$0.001} & 0.143{\scriptsize$\pm$0.002} & 0.135{\scriptsize$\pm$0.001} & 0.130{\scriptsize$\pm$0.002} \\
No delta weighting & 0.112{\scriptsize$\pm$0.000} & 0.100{\scriptsize$\pm$0.001} & 0.095{\scriptsize$\pm$0.001} & 0.132{\scriptsize$\pm$0.001} & 0.127{\scriptsize$\pm$0.001} & 0.123{\scriptsize$\pm$0.001} & 0.149{\scriptsize$\pm$0.001} & 0.138{\scriptsize$\pm$0.001} & 0.133{\scriptsize$\pm$0.001} \\
No $k$-means++ & 0.111{\scriptsize$\pm$0.001} & 0.101{\scriptsize$\pm$0.001} & 0.096{\scriptsize$\pm$0.001} & \textbf{0.130{\scriptsize$\pm$0.001}} & 0.126{\scriptsize$\pm$0.001} & 0.124{\scriptsize$\pm$0.000} & 0.144{\scriptsize$\pm$0.002} & 0.136{\scriptsize$\pm$0.001} & 0.132{\scriptsize$\pm$0.002} \\
No final embeddings & 0.113{\scriptsize$\pm$0.001} & 0.101{\scriptsize$\pm$0.001} & 0.096{\scriptsize$\pm$0.000} & 0.132{\scriptsize$\pm$0.001} & 0.127{\scriptsize$\pm$0.001} & 0.127{\scriptsize$\pm$0.001} & \textbf{0.142{\scriptsize$\pm$0.001}} & 0.136{\scriptsize$\pm$0.001} & 0.131{\scriptsize$\pm$0.001} \\
$k$-means++ cold start & 0.112{\scriptsize$\pm$0.001} & 0.102{\scriptsize$\pm$0.001} & 0.096{\scriptsize$\pm$0.001} & 0.133{\scriptsize$\pm$0.002} & 0.128{\scriptsize$\pm$0.001} & 0.126{\scriptsize$\pm$0.001} & 0.146{\scriptsize$\pm$0.002} & 0.137{\scriptsize$\pm$0.002} & 0.132{\scriptsize$\pm$0.001} \\
Random cold start & 0.112{\scriptsize$\pm$0.001} & 0.104{\scriptsize$\pm$0.001} & 0.099{\scriptsize$\pm$0.001} & 0.136{\scriptsize$\pm$0.002} & 0.130{\scriptsize$\pm$0.001} & 0.126{\scriptsize$\pm$0.002} & 0.145{\scriptsize$\pm$0.002} & 0.137{\scriptsize$\pm$0.001} & 0.133{\scriptsize$\pm$0.001} \\
Uncertainty weighting & \textbf{0.108{\scriptsize$\pm$0.000}} & \textbf{0.097{\scriptsize$\pm$0.000}} & \textbf{0.091{\scriptsize$\pm$0.000}} & 0.131{\scriptsize$\pm$0.000} & \textbf{0.125{\scriptsize$\pm$0.001}} & \textbf{0.122{\scriptsize$\pm$0.000}} & 0.143{\scriptsize$\pm$0.001} & \textbf{0.133{\scriptsize$\pm$0.001}} & \textbf{0.128{\scriptsize$\pm$0.001}} \\
Top-$k$ delta & 0.122{\scriptsize$\pm$0.001} & 0.110{\scriptsize$\pm$0.001} & 0.103{\scriptsize$\pm$0.001} & 0.141{\scriptsize$\pm$0.002} & 0.136{\scriptsize$\pm$0.002} & 0.136{\scriptsize$\pm$0.000} & 0.164{\scriptsize$\pm$0.005} & 0.153{\scriptsize$\pm$0.005} & 0.155{\scriptsize$\pm$0.006} \\
Top-$k$ uncertainty & 0.121{\scriptsize$\pm$0.002} & 0.107{\scriptsize$\pm$0.001} & 0.100{\scriptsize$\pm$0.001} & 0.136{\scriptsize$\pm$0.001} & 0.129{\scriptsize$\pm$0.001} & 0.126{\scriptsize$\pm$0.001} & 0.153{\scriptsize$\pm$0.002} & 0.145{\scriptsize$\pm$0.001} & 0.139{\scriptsize$\pm$0.002} \\
\midrule
\multicolumn{10}{l}{\emph{Pearson --- 1 adaptive round}} \\
\midrule
Proposed & 0.844{\scriptsize$\pm$0.004} & 0.866{\scriptsize$\pm$0.003} & 0.875{\scriptsize$\pm$0.002} & 0.700{\scriptsize$\pm$0.003} & 0.723{\scriptsize$\pm$0.004} & 0.730{\scriptsize$\pm$0.002} & 0.673{\scriptsize$\pm$0.015} & 0.714{\scriptsize$\pm$0.005} & 0.727{\scriptsize$\pm$0.006} \\
No delta weighting & 0.844{\scriptsize$\pm$0.003} & 0.867{\scriptsize$\pm$0.002} & 0.878{\scriptsize$\pm$0.002} & 0.686{\scriptsize$\pm$0.006} & 0.712{\scriptsize$\pm$0.005} & 0.725{\scriptsize$\pm$0.003} & 0.662{\scriptsize$\pm$0.010} & 0.703{\scriptsize$\pm$0.004} & 0.721{\scriptsize$\pm$0.003} \\
No $k$-means++ & 0.839{\scriptsize$\pm$0.003} & 0.861{\scriptsize$\pm$0.003} & 0.872{\scriptsize$\pm$0.002} & \textbf{0.701{\scriptsize$\pm$0.002}} & 0.722{\scriptsize$\pm$0.003} & 0.727{\scriptsize$\pm$0.003} & 0.676{\scriptsize$\pm$0.014} & 0.710{\scriptsize$\pm$0.006} & 0.725{\scriptsize$\pm$0.006} \\
No final embeddings & 0.838{\scriptsize$\pm$0.004} & 0.861{\scriptsize$\pm$0.003} & 0.872{\scriptsize$\pm$0.001} & 0.696{\scriptsize$\pm$0.004} & 0.715{\scriptsize$\pm$0.003} & 0.725{\scriptsize$\pm$0.002} & 0.676{\scriptsize$\pm$0.011} & 0.711{\scriptsize$\pm$0.006} & 0.725{\scriptsize$\pm$0.006} \\
$k$-means++ cold start & 0.842{\scriptsize$\pm$0.003} & 0.861{\scriptsize$\pm$0.003} & 0.872{\scriptsize$\pm$0.001} & 0.696{\scriptsize$\pm$0.008} & 0.717{\scriptsize$\pm$0.006} & 0.728{\scriptsize$\pm$0.002} & 0.669{\scriptsize$\pm$0.011} & 0.706{\scriptsize$\pm$0.004} & 0.721{\scriptsize$\pm$0.006} \\
Random cold start & 0.842{\scriptsize$\pm$0.004} & 0.858{\scriptsize$\pm$0.003} & 0.866{\scriptsize$\pm$0.003} & 0.682{\scriptsize$\pm$0.008} & 0.709{\scriptsize$\pm$0.004} & 0.721{\scriptsize$\pm$0.004} & 0.665{\scriptsize$\pm$0.010} & 0.695{\scriptsize$\pm$0.010} & 0.718{\scriptsize$\pm$0.009} \\
Uncertainty weighting & \textbf{0.850{\scriptsize$\pm$0.001}} & \textbf{0.873{\scriptsize$\pm$0.001}} & \textbf{0.884{\scriptsize$\pm$0.001}} & 0.698{\scriptsize$\pm$0.003} & \textbf{0.723{\scriptsize$\pm$0.002}} & \textbf{0.734{\scriptsize$\pm$0.001}} & \textbf{0.681{\scriptsize$\pm$0.004}} & \textbf{0.720{\scriptsize$\pm$0.002}} & \textbf{0.734{\scriptsize$\pm$0.004}} \\
Top-$k$ delta & 0.816{\scriptsize$\pm$0.002} & 0.844{\scriptsize$\pm$0.001} & 0.858{\scriptsize$\pm$0.001} & 0.654{\scriptsize$\pm$0.005} & 0.689{\scriptsize$\pm$0.004} & 0.698{\scriptsize$\pm$0.002} & 0.598{\scriptsize$\pm$0.013} & 0.654{\scriptsize$\pm$0.003} & 0.666{\scriptsize$\pm$0.006} \\
Top-$k$ uncertainty & 0.822{\scriptsize$\pm$0.003} & 0.852{\scriptsize$\pm$0.002} & 0.867{\scriptsize$\pm$0.002} & 0.667{\scriptsize$\pm$0.006} & 0.700{\scriptsize$\pm$0.005} & 0.713{\scriptsize$\pm$0.004} & 0.628{\scriptsize$\pm$0.006} & 0.661{\scriptsize$\pm$0.012} & 0.675{\scriptsize$\pm$0.012} \\
\midrule
\multicolumn{10}{l}{\emph{Pearson --- 2 adaptive rounds}} \\
\midrule
Proposed & 0.848{\scriptsize$\pm$0.004} & 0.867{\scriptsize$\pm$0.003} & 0.877{\scriptsize$\pm$0.002} & 0.701{\scriptsize$\pm$0.006} & 0.724{\scriptsize$\pm$0.003} & 0.732{\scriptsize$\pm$0.001} & 0.679{\scriptsize$\pm$0.012} & 0.714{\scriptsize$\pm$0.004} & 0.731{\scriptsize$\pm$0.009} \\
No delta weighting & 0.843{\scriptsize$\pm$0.002} & 0.866{\scriptsize$\pm$0.002} & 0.878{\scriptsize$\pm$0.003} & 0.691{\scriptsize$\pm$0.006} & 0.710{\scriptsize$\pm$0.002} & 0.727{\scriptsize$\pm$0.003} & 0.653{\scriptsize$\pm$0.008} & 0.698{\scriptsize$\pm$0.007} & 0.720{\scriptsize$\pm$0.005} \\
No $k$-means++ & 0.841{\scriptsize$\pm$0.004} & 0.864{\scriptsize$\pm$0.003} & 0.873{\scriptsize$\pm$0.002} & \textbf{0.705{\scriptsize$\pm$0.002}} & \textbf{0.724{\scriptsize$\pm$0.001}} & 0.731{\scriptsize$\pm$0.002} & 0.677{\scriptsize$\pm$0.008} & 0.715{\scriptsize$\pm$0.004} & 0.729{\scriptsize$\pm$0.006} \\
No final embeddings & 0.839{\scriptsize$\pm$0.005} & 0.863{\scriptsize$\pm$0.003} & 0.872{\scriptsize$\pm$0.002} & 0.701{\scriptsize$\pm$0.004} & 0.722{\scriptsize$\pm$0.003} & 0.729{\scriptsize$\pm$0.003} & \textbf{0.683{\scriptsize$\pm$0.012}} & 0.711{\scriptsize$\pm$0.005} & 0.729{\scriptsize$\pm$0.007} \\
$k$-means++ cold start & 0.844{\scriptsize$\pm$0.005} & 0.862{\scriptsize$\pm$0.003} & 0.874{\scriptsize$\pm$0.003} & 0.697{\scriptsize$\pm$0.005} & 0.717{\scriptsize$\pm$0.005} & 0.728{\scriptsize$\pm$0.004} & 0.668{\scriptsize$\pm$0.011} & 0.707{\scriptsize$\pm$0.006} & 0.725{\scriptsize$\pm$0.005} \\
Random cold start & 0.845{\scriptsize$\pm$0.004} & 0.859{\scriptsize$\pm$0.004} & 0.867{\scriptsize$\pm$0.003} & 0.682{\scriptsize$\pm$0.009} & 0.708{\scriptsize$\pm$0.004} & 0.723{\scriptsize$\pm$0.004} & 0.670{\scriptsize$\pm$0.011} & 0.702{\scriptsize$\pm$0.009} & 0.720{\scriptsize$\pm$0.008} \\
Uncertainty weighting & \textbf{0.851{\scriptsize$\pm$0.001}} & \textbf{0.873{\scriptsize$\pm$0.001}} & \textbf{0.887{\scriptsize$\pm$0.001}} & 0.697{\scriptsize$\pm$0.003} & 0.721{\scriptsize$\pm$0.005} & \textbf{0.733{\scriptsize$\pm$0.003}} & 0.679{\scriptsize$\pm$0.005} & \textbf{0.716{\scriptsize$\pm$0.005}} & \textbf{0.737{\scriptsize$\pm$0.004}} \\
Top-$k$ delta & 0.821{\scriptsize$\pm$0.003} & 0.848{\scriptsize$\pm$0.003} & 0.860{\scriptsize$\pm$0.002} & 0.666{\scriptsize$\pm$0.003} & 0.695{\scriptsize$\pm$0.004} & 0.701{\scriptsize$\pm$0.002} & 0.606{\scriptsize$\pm$0.016} & 0.657{\scriptsize$\pm$0.003} & 0.672{\scriptsize$\pm$0.012} \\
Top-$k$ uncertainty & 0.823{\scriptsize$\pm$0.004} & 0.854{\scriptsize$\pm$0.002} & 0.870{\scriptsize$\pm$0.001} & 0.678{\scriptsize$\pm$0.008} & 0.709{\scriptsize$\pm$0.005} & 0.718{\scriptsize$\pm$0.002} & 0.643{\scriptsize$\pm$0.008} & 0.674{\scriptsize$\pm$0.006} & 0.699{\scriptsize$\pm$0.007} \\
\bottomrule
\end{tabular}
}
\end{table}

\paragraph{Delta weighting is the most important component.}
Removing delta weighting (``No delta weighting'') consistently degrades hit recall across all datasets and budgets, with losses of 2--6 percentage points on BATCHIE and 2--3 on GDSC2.
This confirms that prioritizing treatments with large predicted differential sensitivity is essential for hit detection.

\paragraph{Diversity is essential.}
Pure exploitation strategies (Top-$k$ delta, Top-$k$ uncertainty) that select the most promising treatments without diversity constraints perform substantially worse than their diversity-aware counterparts across all metrics.
On NCI-ALMANAC, Top-$k$ delta achieves only 51.1\% hit recall at budget 300, compared to 62.9\% for the proposed method, a loss of nearly 12 points.
This demonstrates that greedy selection leads to redundant observations that fail to improve the model's global predictive accuracy.

\paragraph{$k$-means++ outperforms $k$-medoids for adaptive selection.}
Replacing $k$-means++ with $k$-medoids in the adaptive rounds (``No $k$-means++'') reduces hit recall by up to 4 points on BATCHIE, though the difference is smaller on GDSC2 and NCI-ALMANAC.

\paragraph{Uncertainty weighting improves prediction but not hit detection.}
Weighting by MC dropout uncertainty instead of predicted delta (``Uncertainty weighting'') achieves the best MAE and Pearson correlation on all three datasets, indicating that uncertainty-guided selection improves overall predictive accuracy.
However, it underperforms delta weighting for hit recall by 4--6 points on BATCHIE, consistent with the observation that uncertainty targets regions of model ignorance rather than regions of differential drug sensitivity.

\paragraph{Cold-start method matters less when followed by adaptive rounds.}
When only 33\% of the budget is allocated to the initial batch and adaptive rounds follow, the cold-start method ($k$-medoids, $k$-means++, or random) has limited impact on the final hit recall, as the adaptive rounds compensate for suboptimal initial coverage.
However, when the full budget is allocated to the initial batch (i.e., no adaptive rounds), $k$-medoids on drug-dose embeddings substantially outperforms random selection (\cref{tab:al_hit_recall}).

\paragraph{Embedding choice for adaptive rounds.}
Using drug-dose embeddings instead of final model embeddings for adaptive selection (``No final embeddings'') yields mixed results: slightly worse on BATCHIE but comparable or better on GDSC-SQ and NCI-ALMANAC, suggesting that the benefit of response-conditioned embeddings is dataset-dependent.

% ======================================================================
\newpage

% \clearpage
% \input{sections/checklist}

\end{document}